\documentclass[journal]{IEEEtran}
\usepackage[caption=false,font=footnotesize]{subfig}
\usepackage{color}
\usepackage{xcolor}
\usepackage{cite}
\usepackage{amsmath}
\usepackage{amssymb}
\usepackage{amsthm}
\usepackage{graphicx}

\usepackage{epstopdf}
\usepackage{comment}
\usepackage{multirow}

\usepackage{paralist}
\usepackage{lineno}
\usepackage{mdwlist}
\usepackage{eurosym}\DeclareGraphicsExtensions{.pdf,.png,.jpg}
\usepackage{breqn}
\usepackage{makecell}
\usepackage{soul} 
\graphicspath{{./pic/}}

\usepackage{subfig}
\usepackage{bm}
\usepackage[normalem]{ulem}
\usepackage{ulem}

\newtheorem{remark}{\bfseries Remark}
\theoremstyle{remark}

\begin{document}
	\title{Wind Power Scenario Generation based on the Generalized Dynamic Factor Model and Generative Adversarial Network}
	\author{
		Young-ho~Cho,~\IEEEmembership{Student Member,~IEEE,}
        Hao~Zhu,~\IEEEmembership{Senior Member,~IEEE,}
		Duehee~Lee,~\IEEEmembership{Member,~IEEE,}\\
		and~Ross Baldick,~\IEEEmembership{Fellow,~IEEE}}
	
	\markboth{IEEE Transactions on Power Systems}
	{Shell \MakeLowercase{\textit{et al.}}: Bare Demo of IEEEtran.cls for Journals}
	\maketitle

\begin{abstract}
For conducting resource adequacy studies, we synthesize multiple long-term wind power scenarios of distributed wind farms simultaneously by using the spatio-temporal features: spatial and temporal correlation, waveforms, marginal and ramp rates distributions of waveform, power spectral densities, and statistical characteristics. 
Generating the spatial correlation in scenarios requires the design of common factors for neighboring wind farms and antithetical factors for distant wind farms. The generalized dynamic factor model (GDFM) can extract the common factors through cross spectral density analysis, but it cannot closely imitate waveforms.
The GAN can synthesize plausible samples representing the temporal correlation by verifying samples through a fake sample discriminator.
To combine the advantages of GDFM and GAN, we use the GAN to provide a filter that extracts dynamic factors with temporal information from the observation data, and we then apply this filter in the GDFM to represent both spatial and frequency correlations of plausible waveforms. Numerical tests on the combination of GDFM and GAN have demonstrated performance improvements over competing alternatives in synthesizing wind power scenarios from Australia, better realizing plausible statistical characteristics of actual wind power compared to alternatives such as the GDFM with a filter synthesized from distributions of actual dynamic filters and the GAN with direct synthesis without dynamic factors.
\end{abstract}


\section{Introduction}
\IEEEPARstart{R}{esource} adequacy means to maintain power system reliability by having sufficient capacity such that, even with failures or variability of resources, the probability of not being able to meet all load is sufficiently small~\cite{Resource}. System operators achieve resource adequacy of a power system by ensuring there is enough generation capacity~\cite{cao2021risk}. In the case of intermittent energy resources, the effective load carrying capacity (ELCC) of the intermittent resource is the equivalent capacity of highly reliable generators that would result in the same probability of not being able to meet all load~\cite{HongChen}. For example, the ELCC of wind power can be obtained by simulating power systems with long-term wind power scenarios with realistic ramping rates and marginal distributions~\cite{Guanchi}. Furthermore, the capacity factor and reserve margin contribution of wind power to the power system reliability can also be obtained by simulating a future power system by using realistic long-term wind power scenarios~\cite{billinton2012adequacy}.

To calculate the ELCC of wind power and reserve margin of future power system with high penetration level of wind power, we should develop an algorithm that can synthesize as many long-term wind power scenarios as desired. Although we can calculate the ELCC by analyzing correlation and variability of historical wind data, when the duration of wind power is limited, and when not all combinations of observed characteristics are exhibited in the data~\cite{apt2007spectrum}, we have to synthesize long-term wind power scenarios~\cite{lee2018bivariate}.

In order to be realistic, scenarios should match several long-term statistical characteristics of actual wind power. First, wind power scenarios should have the same marginal distributions as actual wind power, which represent possible deliverable amounts of energy~\cite{zhang2013modeling}. In addition, scenarios should have temporal characteristics so that they can have similar ramping events~\cite{cao2019probabilistic}. For distributed wind farms, scenarios should be spatially and temporally correlated with each other depending on locations~\cite{wang2017probabilistic}. For long-term forecasting, the normalized scenarios should be synthesized to be scaled up to a future expected wind power capacity~\cite{carreno2021long}.

There have been many studies to model long-term statistical characteristics of wind power. 
The distribution of wind power was modeled to assess the potential amount of wind power at the given number of wind turbines in~\cite{ackermann2012wind}. However, better stochastic characteristics could be obtained if the distribution of ramping events were represented as a function of ramping intervals. 
In~\cite{lee2013future}, the trend of spectral density of total wind power was modeled to assess the variability of future total wind power as more wind farms are aggregated. The relative sizes of ramping events of scenarios decrease as more wind farms are aggregated. However, if the spectral densities of the cross correlation among geographically spread wind farms were used, spatially and temporally correlated wind farms could be modeled while representing the decreasing relative sizes of ramping events under aggregation. 
In \cite{li2021privacy}, the spatial and temporal correlation between pairs of wind farms was modeled based on a static Pearson correlation coefficient. However, the correlation could be better modeled if time-varying correlation coefficients among wind farms were measured. Therefore, these characteristics of wind power should be further modeled through advanced statistical models in order to be reflected into wind power scenarios.

Many traditional statistical models have been used to synthesize wind power scenarios. For example, \cite{ma2013scenario} generated scenarios from a multivariate normal distribution; however, representing the complex distribution of ramping events is challenging when only marginal distributions are considered. Similarly, \cite{morales2010methodology} utilized an autoregressive moving average (ARMA) approach to synthesize wind speed scenarios and then convert them into wind power scenarios that capture correlated ramp events, though ARMA models are generally limited to short-term movements. While methods such as applying Fourier transforms can improve the design of scenarios by emphasizing amplitude and duration, they remain constrained by the underlying distributional assumptions.

An alternative advanced statistical method is the generalized dynamic factor model (GDFM), which constructs multivariate time-series models by extracting dynamic factors from the frequency-domain cross-correlations among multiple wind power signals \cite{forni2000generalized}. In the GDFM framework, wind power waveforms are decomposed into a common component—representing concurrent movements across farms—and an idiosyncratic component—capturing site-specific variations. Although \cite{lee2016load} demonstrated that the GDFM can effectively synthesize load and wind power scenarios for applications such as transmission planning, its nonmodular design limits the generation of diverse samples and the representation of marginal distributions without additional sampling strategies.

Recently, generative adversarial networks (GANs) have emerged as a powerful nonparametric alternative for scenario generation. Unlike traditional statistical or factor‐based approaches that require an explicit a priori distribution, GANs learn to replicate complex, high-dimensional data distributions directly from historical observations. In a GAN framework, the generator network produces synthetic power output signals while the discriminator network evaluates their authenticity, iteratively refining the outputs through an adversarial process \cite{goodfellow2014generative, arjovsky2017wasserstein}. Several GAN variants have been applied to renewable energy scenario generation. For instance, \cite{chen2018model} synthesized wind and solar power scenarios while incorporating spatial correlations, whereas sequence GANs \cite{liang2019sequence} and deep convolutional GANs (DCGANs) \cite{wang2022approach} have further enhanced the modeling of temporal and spatial dependencies. Moreover, recent advances—such as conditional GANs with style-based enhancements (e.g., conditional StyleGAN2 as reported in \cite{yuan2022conditional})—improve controllability by incorporating external conditions (e.g., weather data and forecast error metrics) into an implicit latent space, all without requiring a preset a priori distribution. Techniques such as the Wasserstein GAN with Gradient Penalty \cite{gulrajani2017improved} and spectral normalization \cite{miyato2018spectral} further stabilize training and mitigate mode collapse. Despite these advances, many existing studies have not adequately captured ramping characteristics, which are critical for accurately estimating ramp sizes, regulation ranges, and curtailment probabilities \cite{peng2024extreme}.

In this paper, we propose a novel algorithm that integrates the strengths of the GDFM with GAN to synthesize long-term wind power scenarios. First, we modularize practical formulations to design the GDFM so that the GDFM can represent spectral characteristics of any given waveform, particularly for wind power. The GDFM can represent spatial and temporal correlations in the frequency domain and ramping distributions, but it is not as effective at synthesizing waveforms and their marginal distributions.
Then, we create a GAN to represent the dynamic filter in the GDFM by learning the time-series of geographically distributed wind farms~\cite{wu2016learning}, and the GDFM can generate new cross power spectral densities through the dynamic filter. Since the GAN is good at mimicking waveforms, and since the dynamic filter contains frequency information, the GAN can supplement the GDFM to synthesize marginal distributions of waveforms. 
By providing the dynamic filter in the GDFM, the combination of the GDFM and GAN can accurately imitate original spectral densities of wind power and maintain the spatial and temporal characteristics of wind power in common components. Finally, we synthesize new long-term wind power scenarios of each wind farm.
By synthesizing scenarios using wind power from Australia, we verify that our algorithm can generate spatially and temporally correlated scenarios. Our contributions are as follows:
\begin{enumerate}
\item We develop a modular approach to GDFM so that it can be used to synthesize other waveforms by changing its structures and their components for different applications.
\item We synthesize wind power scenarios by combining the GAN and the GDFM so that they can represent the spatial and temporal characteristics in both domains.
\item We show that the transformation of a signal to the cross spectral density in the complex domain can guarantee the shape of spectral density of the signal in the real domain.
\end{enumerate}

\section{Generalized Dynamic Factor Model}

\subsection{Spectral Decomposition}
Time-series signals in multiple locations can be decomposed into periodic and stochastic components. We assume that periodic components are the sum of several periodic terms, and that stochastic components randomly follow statistical characteristics. The GDFM decomposes the stochastic component $\bm X$ into the common $\bm \chi$ and idiosyncratic $\bm \xi$ components as
\begin{align}
\bm X=\bm \chi  + \bm \xi .
\end{align}
For wind power, it is assumed that $\bm \chi$ shares concurrent dynamic factors, such as prevailing winds. It is also assumed that dynamic factors are driven by concurrent dynamic shocks, such as global meteorological changes, through the factor loading due to issues, such as geographic features that affect nearby wind farms similarly, but with differences in timing and magnitude. Furthermore, $\bm \xi$ is assumed to be pure random movements affected by local conditions that are uncorrelated from farm to farm.

We build the matrix of wind power $\bm X$ by grouping the vectors of time-series of wind power $\bm x^n \in \mathbb R^{T \times 1}$ for wind farms $n = 1, \dots, N$ into
\begin{align}
\bm X = [\begin{array}{*{10}{c}}{\bm x^1} , {\bm x^2}, \cdots , {\bm x^N} \end{array}].
\end{align}
Then, we define its $k$-lag covariance matrix $\bm \Phi^X(k) \in \mathbb R^{N \times N}$ as
\begin{align}
{\bm \Phi^X(k)} = \mathbb{E}[\bm X^\top_t \bm X_{t-k}], \quad k = -T,\dots,T.
\end{align}
where $X_t$ is the $t$-th row of $X$.
The correlation in $\bm \Phi^X$ can be represented in the frequency domain by transforming ${\bm \Phi^X}$ into the cross power spectral density (CPSD) ${\bm S^X}({\omega _m}) \in \mathbb R^{N \times N}$ through the discrete Fourier transform as
\begin{align}
{\bm S^X}({\omega _m}) = \sum\limits_{k =  - T}^T {{\bm \Phi^X(k)} \cdot {e^{ - jk{\omega _m}}}},
\end{align}
where $m$ represents the order of frequencies, and $\omega$ represents the sampling frequency.

\begin{figure*}[t!]
\centering
\subfloat[GDFM]{\includegraphics[scale=0.5]{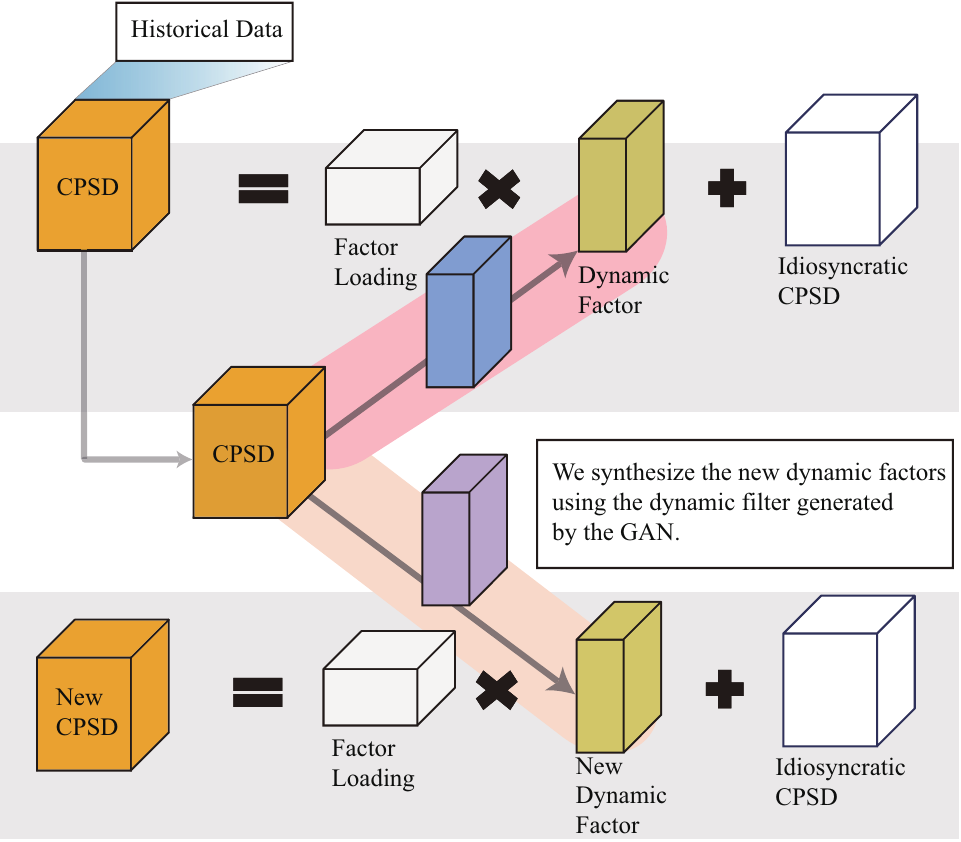}}
\subfloat[GAN]{\includegraphics[scale=0.5]{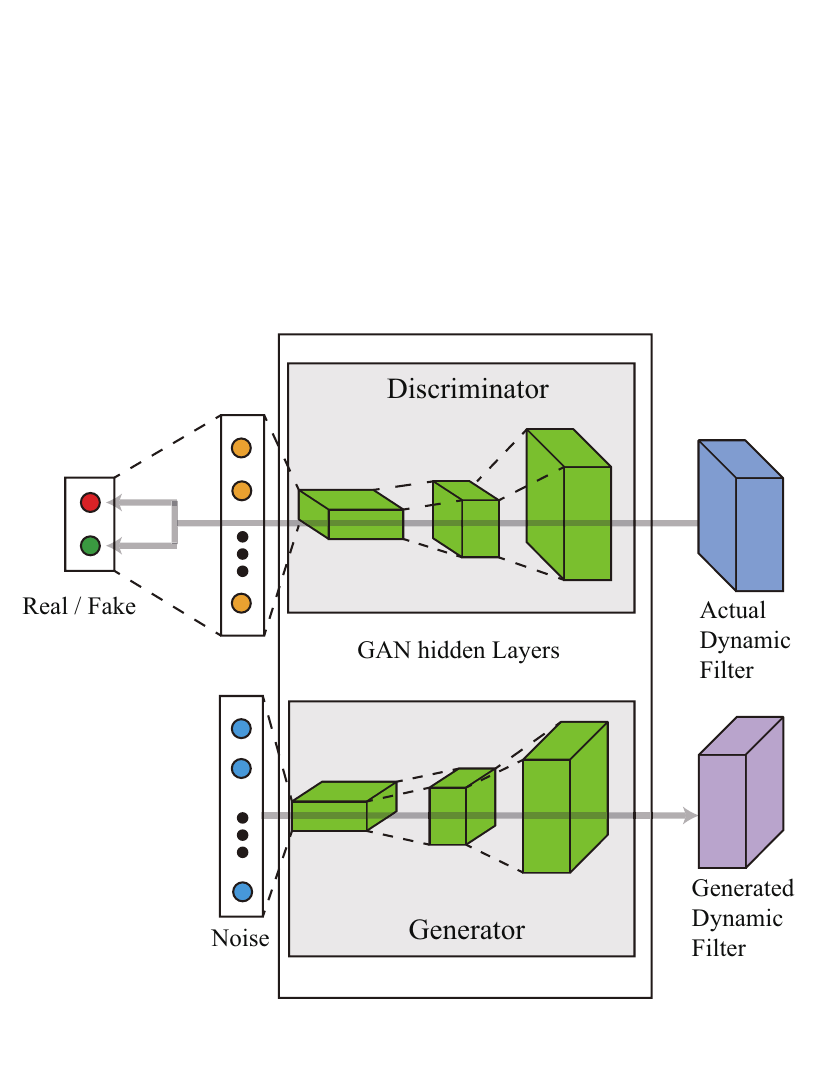}}
\caption{New scenarios are generated through a new CPSD obtained via the (a) GDFM algorithm and the required new dynamic filter is created using (b) GAN. For the case of only using GDFM, the new dynamic filter is synthesized from the distribution of actual dynamic filters. For the case of only using GAN, the time-series data is directly synthesized without modeling each component.} \label{struct}
\end{figure*}

Then, we can decompose $\bm X$ into $\bm \chi$ and $\bm \xi$ in the frequency domain. We decompose $\bm S^X ({\omega _m})$ into the common component in the frequency domain $\bm S^\chi ({\omega _m})$, which is the sum of components having strong intensity, and the idiosyncratic component in the frequency domain $\bm S^\xi ({\omega _m})$, which is the sum of components having weak intensity, as
\begin{align}
\bm S^X({\omega _m}) = \bm S^\chi ({\omega _m}) + \bm S^\xi ({\omega _m}).\label{General}
\end{align}
The split in (\ref{General}) can be accomplished by using the dynamic principal component analysis (DPCA), which is the PCA for each frequency $\omega_m$ in the frequency domain~\cite{forni2000generalized}. We can decompose the measurable variable $\bm S^X ({\omega _m})$ as
\begin{equation}
\label{StartDecomposition}
\bm S^X({\omega _m}) = {\bm V^X }({\omega_m}){\bm \Omega^X }({\omega _m}){\bm V^X }{({\omega _m})^*}, 
\end{equation}
where $^*$ indicates the conjugate transpose matrix, the columns of ${\bm V^X }({\omega_m})$ are the eigenvectors of $\bm S^X ({\omega _m})$, and ${\bm \Omega^X }({\omega _m})$ is the diagonal matrix of eigenvalues of $\bm S^X ({\omega _m})$.
Given an arbitrary choice of the number of $q$, we can further partition ${\bm V^X }({\omega _m})$ into ${\bm V^\chi }({\omega _m})$, which corresponds to the $q$ largest eigenvalues ${\bm \Omega^\chi }({\omega _m})$ based on the magnitudes, and the remaining ${\bm V^\xi }({\omega _m})$ as
\begin{align}
\label{DPCA0}
\bm S^X({\omega _m})&= {\bm V^\chi }({\omega _m}){\bm \Omega^\chi }({\omega _m}){\bm V^\chi }{({\omega _m})^*} \nonumber \\
&\qquad\qquad+{\bm V^\xi }({\omega _m}){\bm \Omega^\xi }({\omega _m}){\bm V^\xi }{({\omega _m})^*}.
\end{align}
The first term of (\ref{DPCA0}) corresponds to $\bm S^\chi ({\omega _m})$, and the second term of (\ref{DPCA0}) corresponds to $\bm S^\xi ({\omega _m})$. By calculating the inverse Fourier Transform of $\bm S^\chi ({\omega _m})$ and $\bm S^\xi ({\omega _m})$, we can calculate $\bm \chi$ and $\bm \xi$, but we need further steps to extract the dynamic factors and their factor loadings.

\subsection{Dynamic Factor Extraction}
If the common components of production at all farms can be explained in terms of $q$ underlying dynamic factors, then we can further decompose $\bm S^\chi ({\omega _ m})$ at the sampling frequency $\omega_m$ into the factor loading $\bm A^\chi({\omega _m}) \in \mathbb R^{N \times q}$ and dynamic factors $\bm f({\omega _m}) \in \mathbb R^{q \times N}$ as
\begin{align}
\label{DynamicFactorLoading}
\bm S^\chi ({\omega _m}) &= \bm A^\chi({\omega _m}) \bm f({\omega _m})
\end{align}
where $\bm A^\chi$ is the coefficient of forward and backward shifts, which we will refer to as the factor loading. It should be noted that the dimension of dynamic factor $q$ is generally smaller than the number of wind farms $N$, so we can represent the data parsimoniously. However, it is difficult to extract $\bm f({\omega _m})$ from $\bm S^\chi ({\omega _m})$ since there exist many possible combinations of $\bm A$ and $\bm f$ satisfying \eqref{DynamicFactorLoading}. We can sufficiently extract a unique $\bm f$ based on three assumptions.

\subsubsection{Filter}
We assume that $\bm f({\omega _m})$ can be obtained by using dynamic filter $\bm B^\chi({\omega _m}) \in \mathbb R^{q \times N}$ from observable $\bm S^X({\omega _m})$ as
\begin{equation}
\label{DynamicFilter}
\bm f({\omega _m}) = \bm B^\chi({\omega _m}) \bm S^X({\omega _m}).
\end{equation}
This assumption allows us to represent $\bm f({\omega _m})$ as the product of $\bm B^\chi({\omega _m})$ and the observed $\bm S^X({\omega _m})$. Then, $\bm S^X({\omega _m})$ can be decomposed as
\begin{equation}
\bm S^X({\omega _m}) = \bm A^\chi({\omega _m}) \bm B^\chi({\omega _m}) \bm S^X ({\omega _m}) + \bm S^\xi ({\omega _m}). \label{General2}
\end{equation}

\subsubsection{Orthogonal Matrix}
$\bm S^\xi ({\omega _m})$ can be represented in the form of
\begin{align}
\bm S^{\xi}({\omega _m}) = \bm A^\xi({\omega _m}) \bm B^\xi({\omega _m}) \bm S^X ({\omega _m}), \label{General3}
\end{align}
where $\bm A({\omega _m})$ and $\bm B({\omega _m})$ should be orthogonal to each other.

\subsubsection{Decomposition of Identity Matrix}
In the DPCA, since the columns of ${\bm V^X }{({\omega _m})}$ corresponding to different eigenvalues are orthogonal to each other, we assume that the identity matrix can be represented as the product of ${\bm V^X }{({\omega _m})}$ and ${\bm V^X }{({\omega _m})^*}$. Then, ${\bm S^X}({\omega _m})$ can be represented as
\begin{equation}
\label{Recursive}
{\bm S^X}({\omega _m}) = {\bm V^X}({\omega _m}){\bm V^X}{({\omega _m})^*}{\bm S^X}({\omega _m}). 
\end{equation}
Since ${\bm V^X}({\omega _m})$ can also be partitioned into ${\bm V^\chi}({\omega _m})$ and ${\bm V^\xi}({\omega _m})$, (\ref{Recursive}) can be decomposed further as
\begin{align}
{\bm S^X}({\omega _m}) &={\bm V^\chi }({\omega _m}){\bm V^\chi }{({\omega _m})^*}{\bm S^X}({\omega _m}) \nonumber \\
&\qquad\qquad+{\bm V^\xi }({\omega _m}){\bm V^\xi }{({\omega _m})^*}{\bm S^X}({\omega _m}).\label{DPCA2}
\end{align}
By comparing (\ref{General2})-(\ref{General3}) and (\ref{DPCA2}), we can define $\bm A({\omega _m})$ as ${\bm V }({\omega _m})$, $\bm B({\omega _m})$ as ${\bm V }({\omega _m})^*$  for both $\bm \chi $ and $\bm \xi$ respectively.

\subsection{Time-Series Extraction in Time Domain}
After we generate dynamic filter $\hat{\bm B}(\omega_m)$ through the GAN, which synthesizes new dynamic factors, we use the same $\bm A^\chi({\omega _m})$, which is an intrinsic characteristic, to synthesize the new CPSD $\hat{\bm S}^\chi({\omega _m})$.
In the following, we will write a $\hat{}$ over a symbol to mean a synthesized value. The magnitude of $\hat{\bm B}(\omega_m)$ is generated by the GAN with the same angles of $\bm B(\omega_m)$. Then, $\hat{\bm \Phi}^\chi(\kappa)$ is obtained by the inverse Fourier transform of $\hat{\bm S}^\chi({\omega _m})$. In this study, we use ${\bm \Phi}^\xi(\kappa)$ from the data to obtain $\hat{\bm \Phi}^X(\kappa)$. 
To allow for the possibility of a synthesized $\Phi^{\xi}$, we will include a $\hat{}$ over the related symbols.

Then, we have to extract new time-series scenarios from $\hat{\bm \Phi}^X(\kappa)$. We assume that there is a zero lag in $\hat{\bm \Phi}^X(\kappa)$. To ignore the leading and lagging effects, we take a zero lag ($\kappa=0$) from $\hat{\bm \Phi}^X(\kappa)$. After taking a zero lag, we can obtain the two-dimensional matrix $\hat{\bm \Phi}^X(0) \in \mathbb R^{N \times N}$. Through the static principal component analysis (SPCA) in~\cite{forni2005generalized}, we decompose $\hat{\bm \Phi}^X(0)$ as
\begin{align}
\hat{\bm \Phi}^X(0) = {\bm P} {\bm \Lambda} {\bm P}',
\end{align}
where $'$ indicates the transpose matrix. We assume $q$ number of static factors, which is the same as the number of dynamic factors. Then, the new wind power $\hat {\bm X}$ is calculated as
\begin{align}
\hat {\bm X} =  \hat{\bm \Phi}^X(0) {\bm P}' ({\bm \Lambda})^{-1} {\bm P} \bm X.
\end{align}
Finally, we can synthesize the new scenario that retains the intrinsic characteristics of the actual scenario.

There are several alternatives for GDFM analysis. For the numerical analysis in Section V, we will use the following variant. Alternatives are described in Section II.D.
In Fig.~\ref{struct}, the scenario generation process is described. Subsets of actual wind power are converted to CPSDs, which are decomposed into CPSDs of common and idiosyncratic components. The CPSD of the common component is decomposed into $\bm A^\chi({\omega _m})$ and $\bm f({\omega _m})$. Then, $\hat{\bm B}(\omega_m)$ is fed to the GAN for training to generate the new $\bm B^\chi({\omega _m})$. The product of ${\bm S^X}({\omega _m})$, $\hat{\bm B}(\omega_m)$, and $\bm A^\chi({\omega _m})$ will be the new $\hat{\bm S}^\chi({\omega _m})$. Finally, after adding $\hat{\bm S}^\xi({\omega _m})$, $\hat{\bm S}^X({\omega _m})$ is obtained and converted to scenarios through the inverse Fourier transform and SPCA. However, scenarios can also be generated by other structures as described in the next section.

\subsection{Structures of GDFM}
\begin{figure}[t!]
\begin{center}
\includegraphics[scale=0.3]{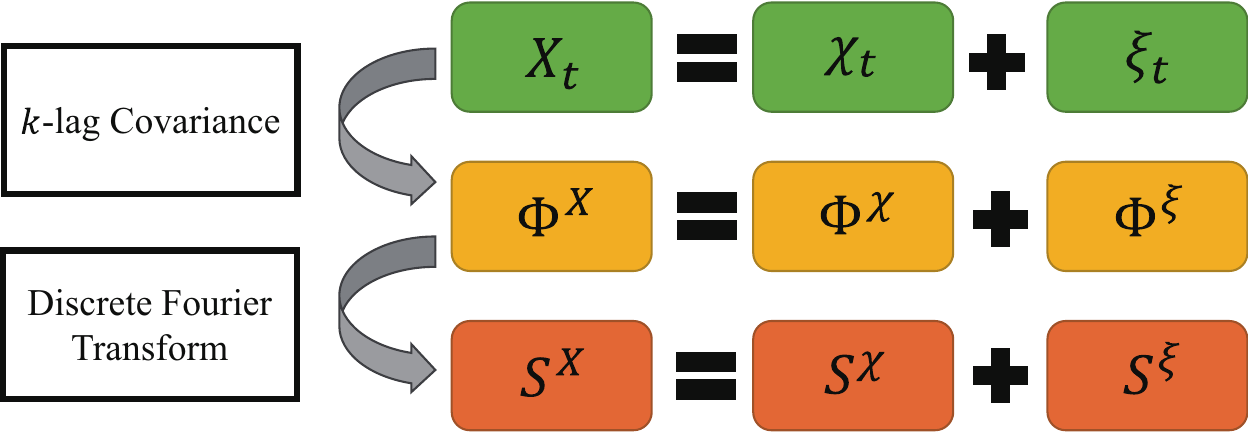}
\caption{The decomposition of the common and idiosyncratic components in the time, covariance, and frequency domains.} \label{ThreeDecomposition}
\end{center}
\end{figure}

We can have several GDFM variants based on three common characteristics of the GDFM. First, the observation data can be decomposed into common and idiosyncratic components in any one of the time, covariance, and frequency domains as shown in Fig.~\ref{ThreeDecomposition}. In the frequency  domain, the decomposition is determined by splitting eigenvalues and the corresponding eigenvectors of spectral densities for the given number of eigenvalues.


Second, there are dynamic factors $\bm f$, which drive the stochastic movement of $\bm \chi$. The GDFM can be classified into two models whether $\bm f$ is defined in the time or frequency domain. If $\bm f$ is defined in the frequency domain as $\bm f({\omega _m})$, $\bm S^\chi ({\omega _m})$ is generated by loading $\bm f({\omega _m})$ through $\bm A^\chi({\omega _m})$ in~(\ref{DynamicFactorLoading}). If $\bm f$ is defined in the time domain as $\bm f_{t}$, $\bm \chi_t$ is generated by loading $\bm f_{t}$ through $\bm A({L})$ as
\begin{equation}
\bm \chi_t =\bm A({L}) \bm f_t.
\end{equation}
Third, there is a filter that can extract $\bm f$ from observable data. In the frequency domain, the dynamic filter $\bm B^\chi({\omega _m})$ extracts $\bm f({\omega _m})$ from $\bm S^X({\omega _m})$ as in (\ref{DynamicFilter}).
Note that this variant is used for the numerical analysis in Section V.
In the time domain, the dynamic filter $\bm B(L)$ extracts $\bm f_{t}$ from $\bm X$ as
\begin{equation}
\label{DynamicFilter2}
\bm f_t  =\bm B(L) \bm X_t.
\end{equation}
It should be noted that the domain of a dynamic factor loading and dynamic filter should be the same.

\begin{table}[t!]
\caption{The formulations of GDFM variants.}
\begin{center}
\begin{tabular}{|c|c|}
\Xhline{3\arrayrulewidth}
Method  & GDFM Formulation \\
\hline
\makecell{Forni's Two-sided \cite{forni2000generalized}} & $\bm X_t=\bm A(L) \bm f_t  + \bm \xi_t$\\
\hline
\makecell{Forni's One-sided \cite{forni2005generalized}} & $\bm X_t=\bm A(L) \bm C(L)^{-1} \bm \epsilon_t   + \bm \xi_t $ \\
\hline
Lee's method \cite{lee2016load}& 
$\bm X_t=\bm A(L) \bm C(L)^{-1} \bm Z \bm \delta_t  + \bm \xi_t $  \\
\hline
This study & $\bm S^X= \bm A({\omega _m}) \bm B({\omega _m}) \bm S^X ({\omega _m}) + \bm S^\xi ({\omega _m})$ \\
\Xhline{3\arrayrulewidth}
\end{tabular} \label{GDFM_compare}
\end{center}
\vspace{-0.2in}
\end{table}

Based on common characteristics, we describe variants in previous studies in Table \ref{GDFM_compare}. First, Forni et al. established the basic structure of the GDFM, known as the two-sided model, in~\cite{forni2000generalized}. In detail, $ \bm f $ was defined in the time domain, but $ \bm f $ was not estimated. Instead of $ \bm f $, $\bm \chi$ was directly estimated from $\bm X$ by using the dynamic filter $\bm K(L)$. 
Furthermore, positive and negative lags were used to load $\bm f_{t}$, so lagged signals were overlapped. This characteristic is applicable to scenario generation, but it is not applicable to real-time forecasting.
%
In \cite{forni2005generalized}, Forni et al. developed the one-sided model, which assumes $\bm f_{t}$ can be further decomposed into dynamic shocks $\bm \epsilon$ and the inverse of noise loading $\bm C(L)$ as
\begin{equation}
 \bm f_t = {\bm C(L)} ^{-1} \epsilon_t,
\end{equation}
where $\bm f_{t}$ is lagged as a time-series of $\bm \epsilon$ to make the GDFM progress forward. However, $\bm f_t$ was not directly estimated.
Finally, in~\cite{lee2016load}, Lee et al. represented $\bm \epsilon$ as loaded random white noise $\bm \delta _t$ on a shock loading $\bm Z$ as
\begin{equation}
\epsilon_t = \bm Z \bm \delta _t ,
\label{LeeLee}
\end{equation}
to make as many scenarios as desired by changing $\bm \delta _t$. By multiplying the inverse of the coefficient matrix of $\bm C(L)$ to $\bm \epsilon$, $\bm f_{t}$ was estimated.

\begin{figure}[t!]
\begin{center}
\includegraphics[scale=0.32]{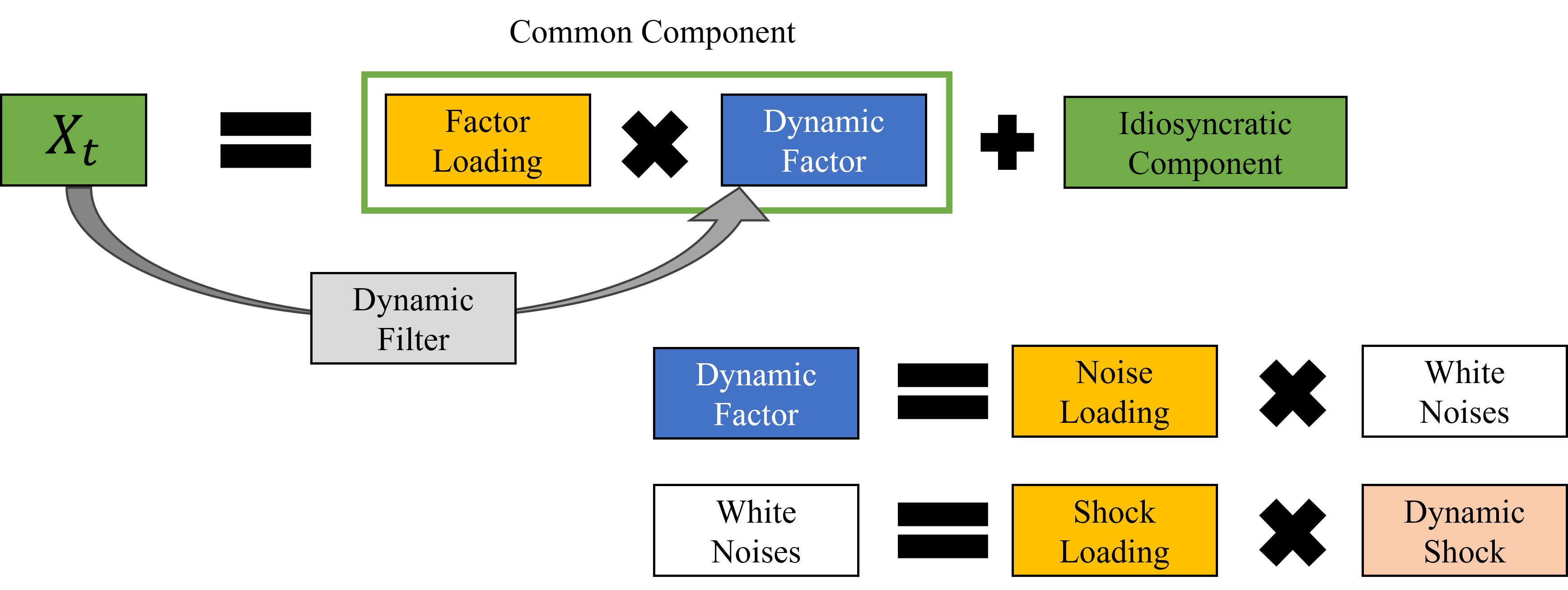}
\caption{The relationships among the elements of the GDFM.} \label{IndividualCompo}
\end{center}
\end{figure}

Based on common characteristics of examples, we can define all elements of the GDFM as follows: common component, idiosyncratic component, dynamic filter, dynamic factor, factor loading, shocks, shock loading, white noise, and noise loading. Their relationships are described in Fig.~\ref{IndividualCompo}. First, the stochastic component of the observation data can be decomposed into the common component and idiosyncratic component. The key assumption is that the common component can be loaded from the dynamic factor through the factor loading.
We can estimate the dynamic factor through the top-down and bottom-up approaches. For the top-down approach, the dynamic factor is filtered from the observation data by using the dynamic filter. We can estimate the dynamic factor by comparing two assumptions that the common component is loaded from the dynamic factor through the factor loading, and that the dynamic factor is filtered from the observation data through the dynamic filter.

For the bottom-up approach, we can assume that the cause of the dynamic factor is the white noise. The dynamic factor can be loaded from white noise through the noise loading. White noise can be loaded from the dynamic shock through the shock loading.
It should be noted that the idiosyncratic component represents the random noise in a local scale. On the contrary, the shock represents the random nature of wind, such as the random start of wind waves or wind ramps in a global scale.

While keeping common characteristics of the GDFM, we can build scenarios by changing GDFM elements. We can choose the domain of dynamic factors, and choose shocks and shock loading. Furthermore, we can determine the number of eigenvalues for the common component.  We can also synthesize random idiosyncratic components and add them to the common components \cite{cho2022similarity}. Moreover, we can synthesize scenarios by modifying the common components. We can also change the dynamic factor $\bm f$, factor loading $\bm A$ or filter $\bm B$ in the frequency or time domain. Therefore, there are many ways to synthesize scenarios from GDFM variants.

\subsection{Suggested Scenario Generation Algorithm}
To synthesize scenarios, we use the GDFM to design dynamic factors in common components by modifying three-dimensional CPSDs among wind farms in the frequency domain since the CPSD can represent the cross-correlation and ramping among wind farms. The core of co-movements in the CPSDs is the dynamic factors, which directs same ramping movements of nearby wind farms, and dynamic factors can be extracted from the CPSD through the dynamic filter. The GAN will provide the dynamic filter to the GDFM.

To synthesize scenarios, we define $\bm f({\omega _m})$ in the frequency domain and synthesize it with the different $\bm B^\chi({\omega _m}) $ and same $\bm A^\chi({\omega _m})$. All processes are performed in the frequency domain. One strong point of our model is that we fit the spectral densities of the filter so that the cumulative energy of the power spectral densities of the wind power follow actual ones. Another strong point is to synthesize different scenarios while maintaining the intrinsic characteristics of wind power in the frequency domain. We assume that the characteristics of wind power are in the coefficients of matrices in the GDFM, such as the $\bm A({\omega _m})$, $\bm B({\omega _m})$, and $\bm f({\omega _m})$. Geographical characteristics are involved in $\bm A({\omega _m})$, and stochastic concurrent movements are involved in $\bm B({\omega _m})$, so we use the same $\bm A({\omega _m})$ but make a new $\bm B({\omega _m})$ by using the GAN. Since the randomness is provided by the generator $G$ in GAN, we can synthesize as many different scenarios as desired.


\section{GDFM-enhanced Generative Adversarial Network}

We describe the generative adversarial network (GAN) and present an approach to using both GDFM and GAN.

\subsection{Generative Adversarial Network}
The goal of the GAN is to generate \emph{fake} data that has similar characteristics to the training data~\cite{liang2019sequence}. In the GAN, the generator $G$ generates the fake data and delivers it to the discriminator $D$. The discriminator $D$ receives the real and fake data randomly, and it classifies whether the received data is real or not. The classification for the fake data is delivered to $G$, and $G$ increases generation performance by reflecting decisions. Since $D$ knows the real data, it can be considered as supervised learning. On the contrary, $G$ does not know the real data and only receives decisions for its own generated data, so it can also be considered as unsupervised learning. It should be noted that a deep neural network is the core of $G$ and $D$.

The GAN increases the performance of fake data generation by competing $G$ and $D$ adversarially. While $D$ maximizes its performance to discriminate between fake and real data, $G$ minimizes the performance of $D$ for the fake data so that $D$ confuses the fake data as real data. Therefore, this process can be represented as a min-max problem, and $G$ and $D$ are updated by solving this problem.

Suppose that $G$ generates fake data $\hat {\bm Y}$ by using random noise $\bm Z$ sampled from the known probability distribution $P_z$, such as the Gaussian or Laplace distributions, as:
\begin{equation}
\hat {\bm Y} = G(\bm Z).
\end{equation}
Although $\hat {\bm Y}$ is fake data generated by the generator from random noise, an ideal generator will make the data appear to the discriminator to be real. To instantiate this, we set the objective of $G$ to be minimized:
\begin{equation}
\label{gen_GAN}
\mathcal{L}_G = \mathbb{E} [\log (1-D(G(Z)))].
\end{equation}
On the other hand, the ideal discriminator will discern the difference between fake data $\hat {\bm Y}$ and real data $\bm Y$. The objective of $D$ is to be maximized:
\begin{align}
\mathcal{L}_D = \mathbb{E} [\log D(Y)] +\mathbb{E} [\log (1-D(G(Z)))]. \label{dis_GAN}
\end{align}
Then, $D$ should maximize $\mathcal{L}_D$ by distinguishing real data from fake data. Finally, the optimal solution is obtained by minimizing $\mathcal{L}_G$ after maximizing $\mathcal{L}_D$ so that the objective function $V(D,G)$ of the GAN can be defined as the mini-max problem, which is the combination of (\ref{gen_GAN}) and (\ref{dis_GAN}) as:
\begin{align}
\min\limits_{G} \max\limits_{D} V(D, G) &= \mathbb{E} [\log D(Y)]\nonumber\\
& \quad + \mathbb{E} [\log (1-D(G(Z)))]. \label{obj_GAN}
\end{align}


While GANs are known to suffer from unstable adversarial training and mode collapse~\cite{shahbazi2022collapse}, they do not require an explicit prior over the data distribution, nor do they incur the heavy computational cost of iterative denoising in diffusion models~\cite{rombach2022high} or the likelihood‐based training of VAEs.  A single sample generation with a GAN requires only one forward pass through the generator~\cite{fang2022diggan}, and training typically converges with far fewer examples than alternative approaches. These properties—no prior specification, lower compute per sample, and reduced data requirements—make GANs particularly well suited for our goal of generating extreme or rare scenarios under tight data constraints.

\subsection{Combination of GDFM and GAN}

We integrate the GDFM with a GAN to synthesize the dynamic filter $\bm B(\omega_m)$ in the frequency domain. In the GDFM framework, $\bm B(\omega_m)$ generates spatially correlated and frequency-correlated dynamic factors $\bm f(\omega_m)$. Consequently, scenarios synthesized using $\bm B(\omega_m)$ inherently retain spectral densities closely resembling those of actual wind power data. Since $\bm B(\omega_m)$ is represented as a matrix of complex values, we decompose it into separate magnitude and phase components. Given that the factor loading $\bm A(\omega_m)$ is the conjugate transpose of $\bm B(\omega_m)$, the original phase information can be retained directly. Thus, the synthesis focuses exclusively on the magnitude component. Formally, the synthesized dynamic filter $\hat{\bm B}(\omega_m)$ is constructed as
\begin{align}
\hat{\bm B}(\omega_m) &= \hat{\bm G}(z)\angle\bm B(\omega_m), \label{GDFM_GAN}
\end{align}
where $\hat{\bm G}(z)$ represents the synthesized magnitude matrix generated by the GAN’s generator $G$, and $\angle\bm B(\omega_m)$ denotes the original phase angles extracted from $\bm B(\omega_m)$.

Unlike conventional GAN-based methods, which typically generate time-series data directly in the time domain, our proposed approach utilizes the GAN (Fig.~\ref{struct}(b)) to synthesize this dynamic filter explicitly within the frequency domain. Specifically, white noise is fed into the transposed convolution layers of $G$ to produce the magnitude matrix $\hat{\bm G}(z)$. This synthesized magnitude is then combined with the previously extracted phase information to form the complete synthesized dynamic filter $\hat{\bm B}(\omega_m)$. Next, both the synthesized filter $\hat{\bm B}(\omega_m)$ and the original filter $\bm B(\omega_m)$, extracted via the GDFM from real wind power data, are input into the discriminator $D$.

The authenticity of the synthesized filter is evaluated by $D$, and its feedback is iteratively passed back to $G$ to progressively improve the realism of the synthesized filters. This adversarial training process continues until convergence is achieved as defined by the GAN objective function in \eqref{obj_GAN}. Upon convergence, the synthesized dynamic filter $\hat{\bm B}(\omega_m)$, accurately capturing spatial and frequency correlations in the frequency domain, is integrated into the GDFM. This integration enables the reconstruction of time-series scenarios that preserve realistic spectral densities, ramping behaviors, and correlation structures—characteristics often inadequately modeled by conventional GAN approaches applied directly in the time domain.


\begin{remark}\label{rmk:GANvariants} (GAN variants.)
\rm Although the present work employs a conventional GAN combined with the GDFM, the underlying concept of combining the frequency-domain modeling capability of the GDFM with a deep generative network is inherently flexible. Our approach leverages the GDFM’s ability in modeling common and idiosyncratic components, which capture spatial and frequency correlations in wind power, with the GAN’s strength in synthesizing realistic marginal distributions and waveforms. In our experiments, even though the GDFM struggles with marginal distribution and waveform reproduction, and the GAN fails to adequately preserve frequency-domain correlations, their combination yields significantly improved scenario realism. This foundational concept is not limited to the specific GAN variant used herein. We posit that any advanced GAN variant (e.g., sequence GANs~\cite{liang2019sequence}, deep convolutional GANs (DCGANs)~\cite{wang2022approach}, conditional StyleGAN2~\cite{yuan2022conditional}, Wasserstein GAN with Gradient Penalty~\cite{gulrajani2017improved}, GNN-based GAN~\cite{cho2023wind}, etc.) can replace the conventional GAN in our framework. In fact, improvements achieved by these GAN variants in training stability and controllability will further enhance performance when combined with the GDFM. The hybrid GDFM+GAN (or GDFM+advanced GAN) model will provide a robust and scalable foundation for wind power scenario generation.
\end{remark}

\begin{figure}[t!]
\begin{center}
\includegraphics[scale=0.25]{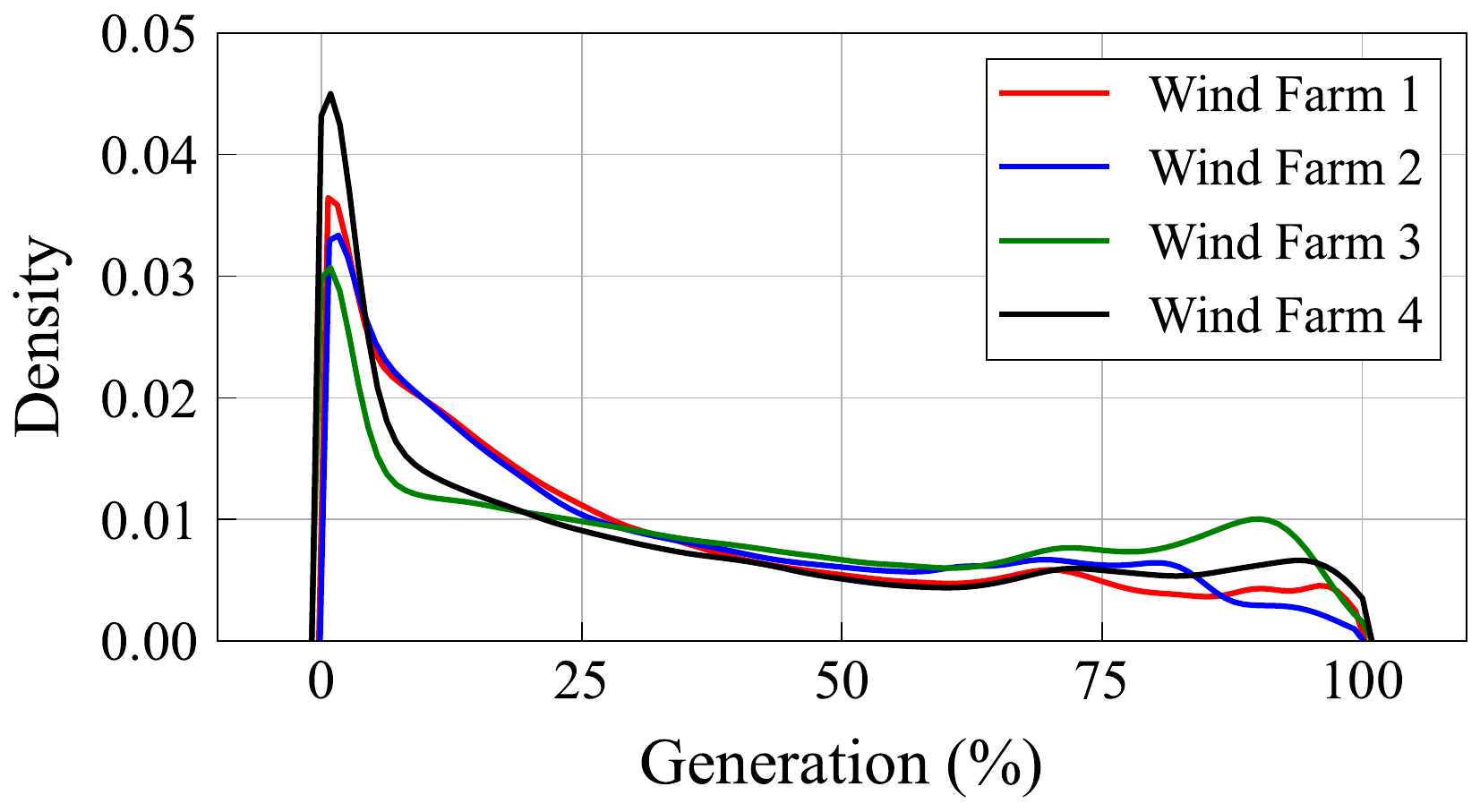}
\caption{Wind power density (WPD) function of historical data for wind farms.} \label{WPD_ex}
\end{center}
\end{figure}

\section{Characteristics of Wind Power Outputs}
In this section, we analyze the characteristics of wind power outputs to verify synthesized scenarios.

\subsection{Variability and Distribution of Ramping Rates}
Fluctuation of wind power can be characterized by its variability and distribution of ramping rates. First, we define the variability as the standard deviation (SD) of ramping rates, which are differences in wind power over a fixed sampling interval. The ramping rates $d_t$ at $t \in {1,..,T}$ is defined for $k$ sampling interval as
\begin{align}
d_t = x_t-x_{t-k},
\end{align}
where $x_t$ is the wind power output. According to~\cite{boutsika2012quantifying}, the distribution of $d_t$ follows the Laplace distribution, which is defined as
\begin{align}
f(x | \mu,b)= \frac{1}{2b} \exp \left(-\frac{|x-\mu|}{b}\right),
\end{align}
where $\mu$ is a location, and $b$ is a scale parameter. We can verify whether the distribution of $d_t$ follows the Laplace distribution by observing, $\mu$, $b$, kurtosis, which measures the tails of the distribution,  and skewness, which measures the asymmetry of the distribution, respectively.

\subsection{Stochastic Movements}
We measure the variability of wind power output by observing the slope of the power spectral density (PSD) in the frequency domain. The PSD is the diagonal elements of the CPSD, which can also be calculated by the discrete Fourier transform of the autocorrelation function. To better measure PSD slopes, we estimate the PSD based on Welch's method \cite{solomon1991psd}. Welch's method mitigates the variance of the PSD estimation by averaging the PSDs so that we can remove fluctuations. In Welch's method, data is divided into $N$ segments, the PSD for each segment is calculated, and the PSDs are averaged. We measure PSD slopes in a logarithmic scale.

\subsection{Capacity Factor}
The capacity factor \emph{CF} of a wind farm can be calculated as
\begin{align}
CF = P/(IC \times 24\text{h} \times 365\text{D}),
\end{align}
where $P$ is the yearly wind power generation and $IC$ is its installed capacity. The empirical probability distributions of power production for four wind farms are in Fig. \ref{WPD_ex}, and $CF$ is calculated based on the distributions. In summary, the distribution of power differences is used to calculate the variability, and the distribution of power itself is used to calculate the \emph{CF}.

\section{Simulation}
We evaluate the performance of combination of the GDFM and GAN (GDFM+GAN) in generating spatially and temporally correlated wind power scenarios.

\subsection{Simulation Settings}

We use wind power outputs sampled every five minutes from 2012 to 2017 from twenty wind farms in Australia~\cite{DATA}. We synthesize three sets of 1,000 scenarios for 100 days to estimate long-term variability by using three models: GDFM, GAN, and GDFM+GAN models.
{\color{blue}Since the GAN can generate only ten days of scenarios at a time—owing to the high memory and computational demands of storing model weights and training data for large systems (e.g., $N\times28800$)—we synthesize 10‐day blocks sequentially and then concatenate them to form 100‐day sequences. 
To validate this concatenation strategy, we first ran experiments at an hourly resolution: we directly synthesized a 100-day scenario and, separately, stitched together ten independent 10-day GAN outputs. The two resulting 100-day series were statistically indistinguishable, confirming that concatenating shorter segments yields valid long-horizon scenarios. Accordingly, for all subsequent simulations at five‐minute resolution, we adopt this concatenation approach.}
We train models by using wind power outputs from 2012 to 2016, and we did not train outputs from 2017 so that we could use data from 2017 to verify the plausibility of our synthesized scenarios.


To synthesize long-term scenarios, our proposed framework (GDFM+GAN) introduces randomness by synthesizing multiple dynamic filters using a GAN and then integrating these filters into the GDFM. In our first comparison case, we use the GDFM alone, where the dynamic filter is sampled directly from the empirical distribution of actual dynamic filters rather than being synthesized. In a second comparison case, we directly apply the GAN to the raw data without integrating it with the GDFM.

Before we synthesize scenarios by using the GDFM, we extract daily trends, which are defined as the average of actual wind power outputs for the given season at every sampling interval for 24 hours, so we have four different seasonal scenarios. Then, we extract seasonal trends from actual wind power outputs at every day. The remaining waveforms are normalized by their capacities. After we synthesize raw scenarios, we add seasonal trends, and we finally apply the mean and standard deviation to have final scenarios.

\begin{figure}[t!]
\begin{center}
\centering
\subfloat[Actual wind power outputs in 2017]{\includegraphics[scale=0.4]{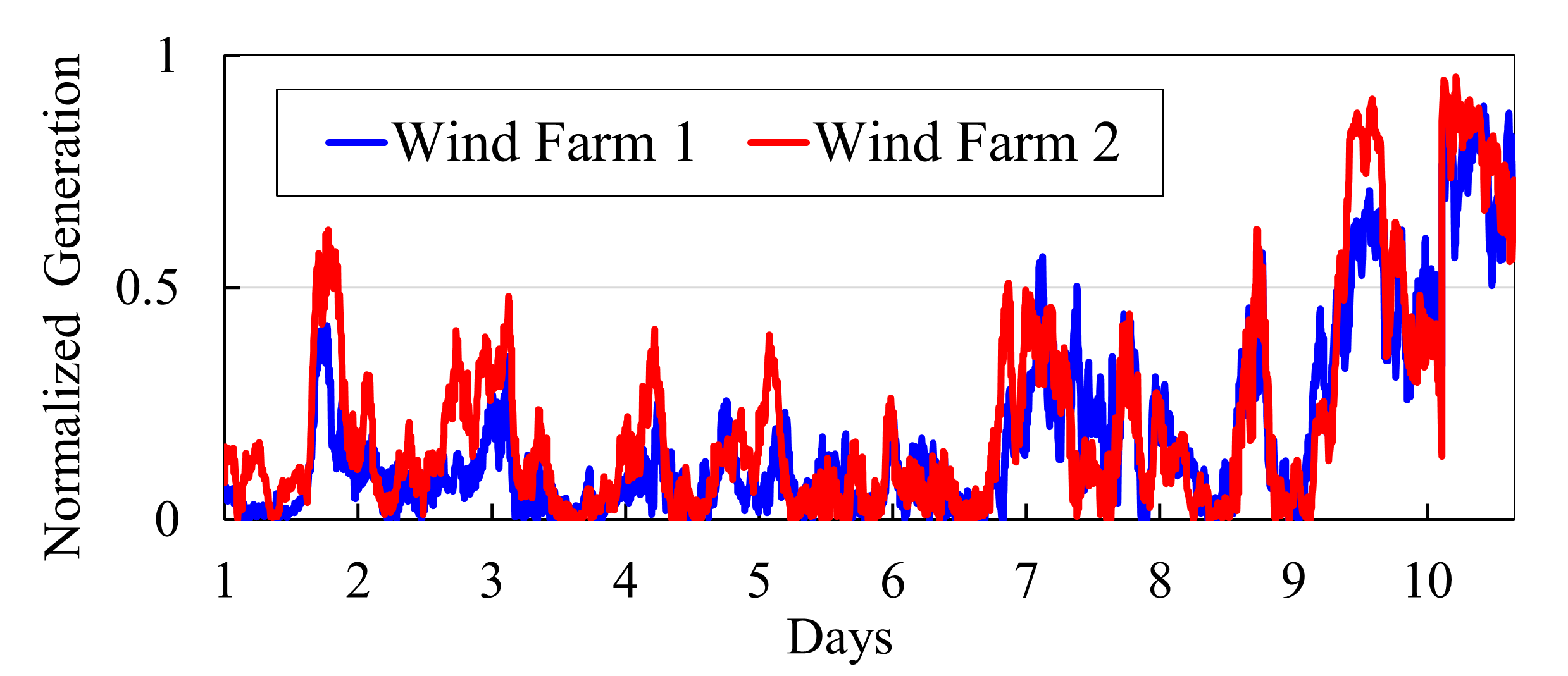}}\\ \vspace{-0.15in}
\subfloat[GDFM]{\includegraphics[scale=0.4]{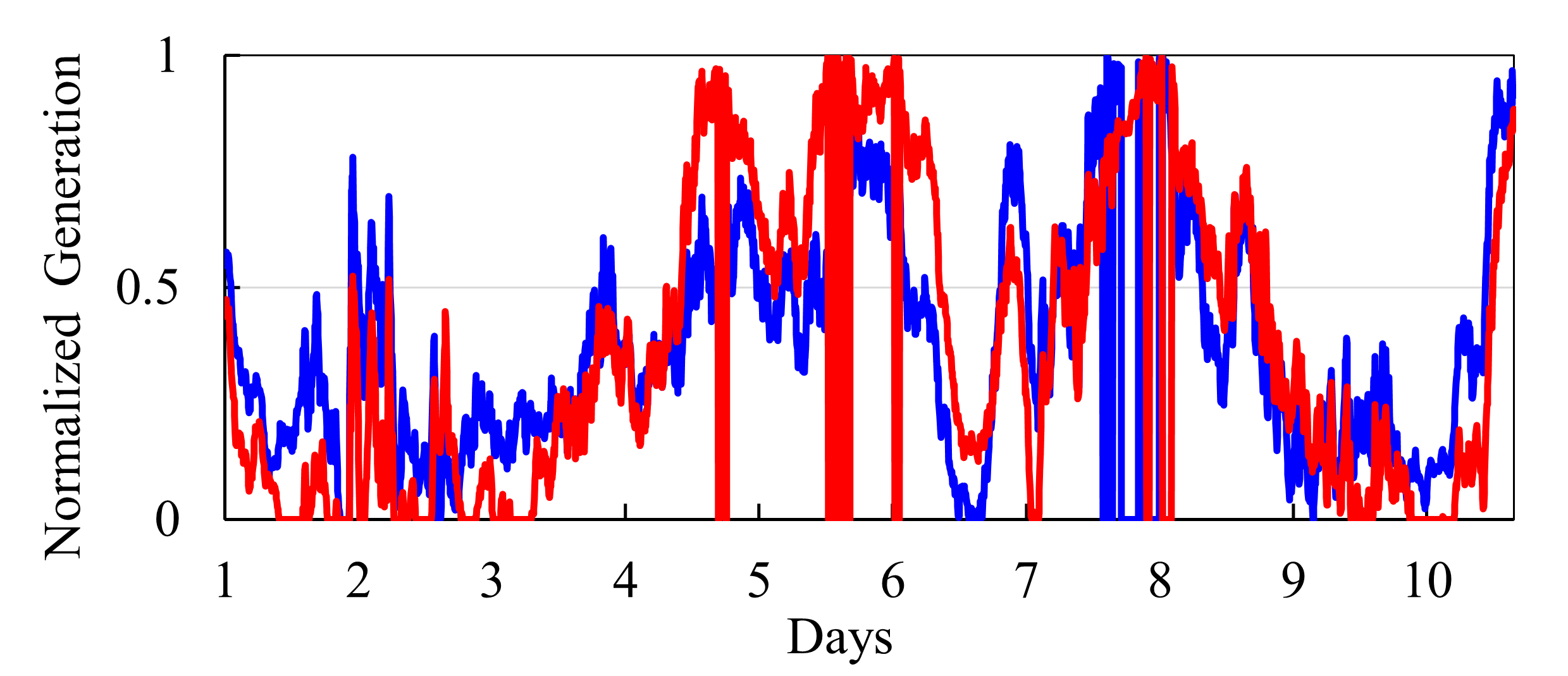}}\\\vspace{-0.15in}
\subfloat[GAN]{\includegraphics[scale=0.4]{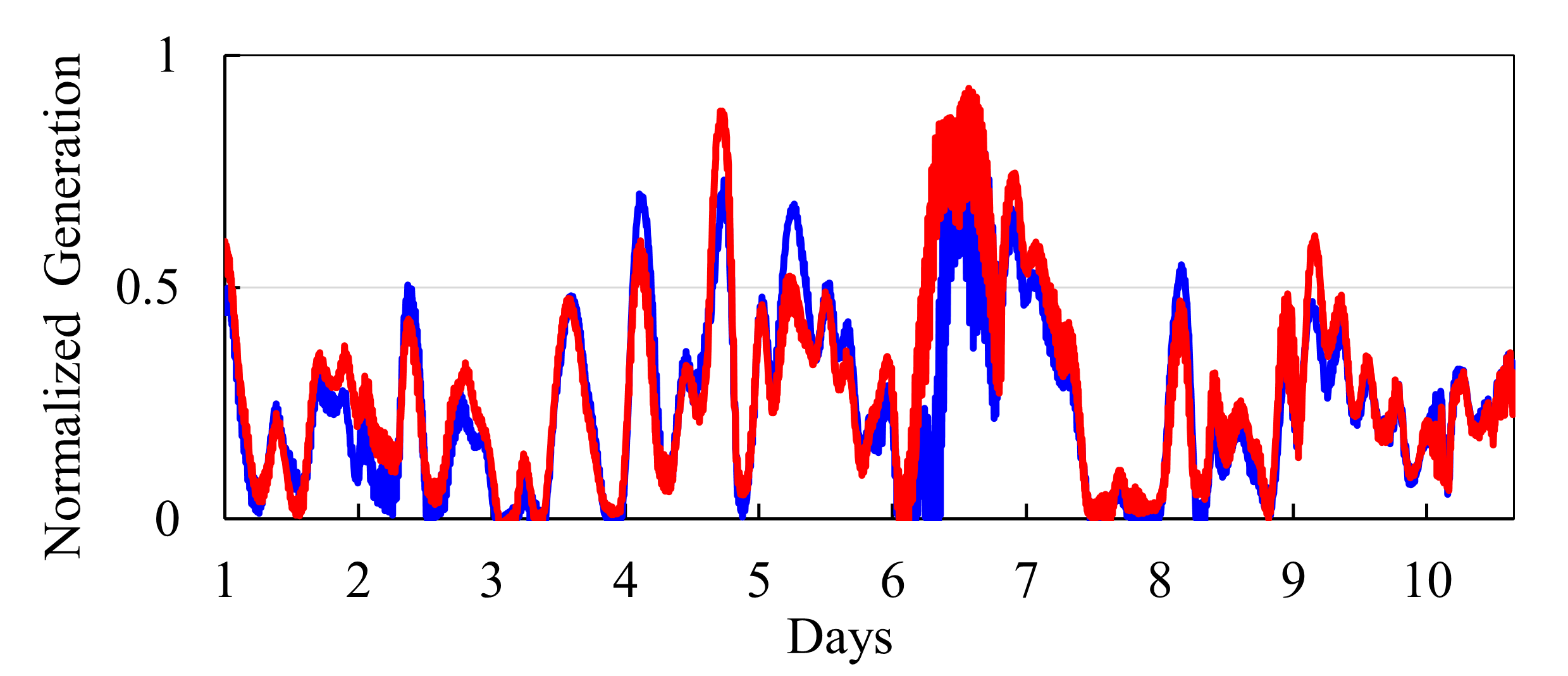}}\\
\vspace{-0.15in}
\subfloat[GDFM+GAN]{\includegraphics[scale=0.4]{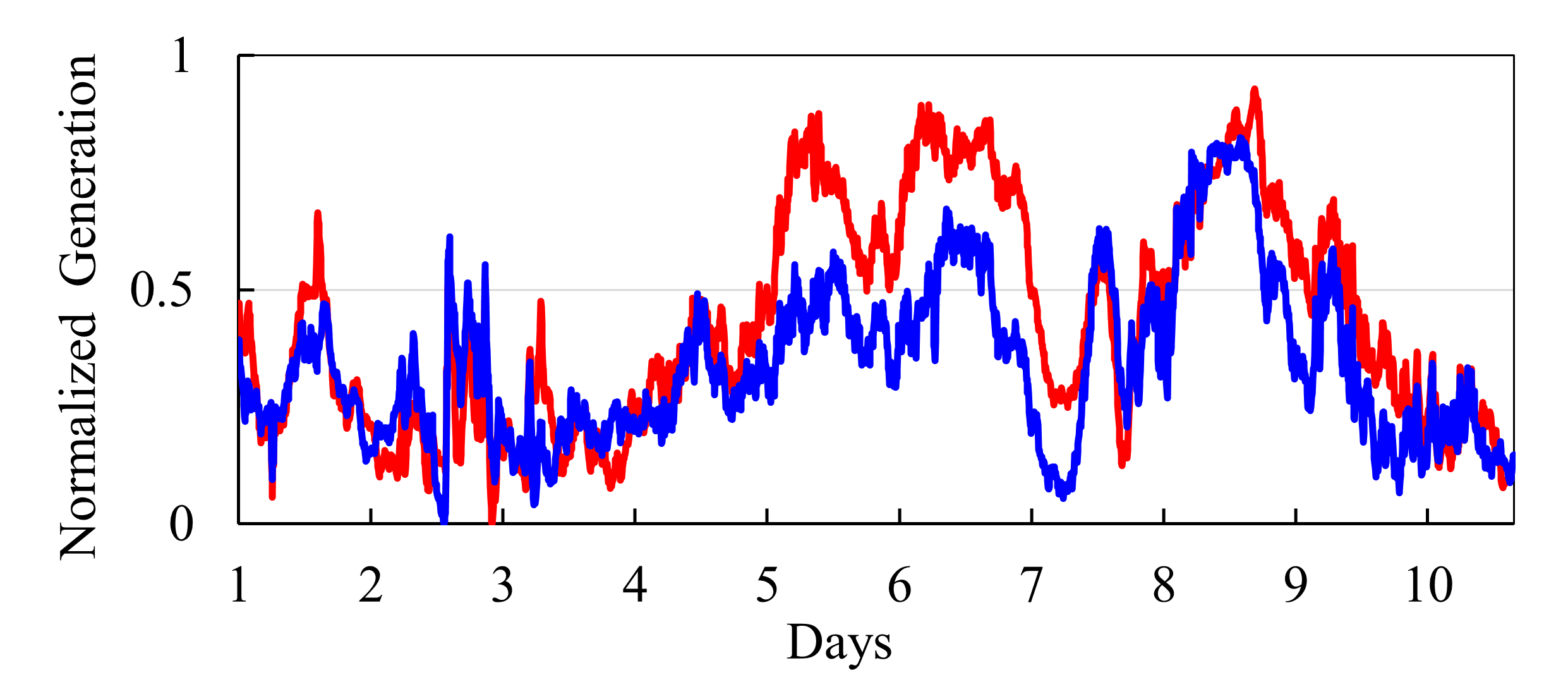}}
\caption{Comparison of 10-day wind power outputs under normal conditions from two adjacent Australian wind farms: (a) actual measurements and scenarios synthesized using (b) GDFM, (c) GAN, and (d) GDFM+GAN.} \label{scenario}
\end{center}
\end{figure}

\begin{figure}[t!]
\begin{center}
\centering
\subfloat[Wind Farm 1]{\includegraphics[scale=0.4]{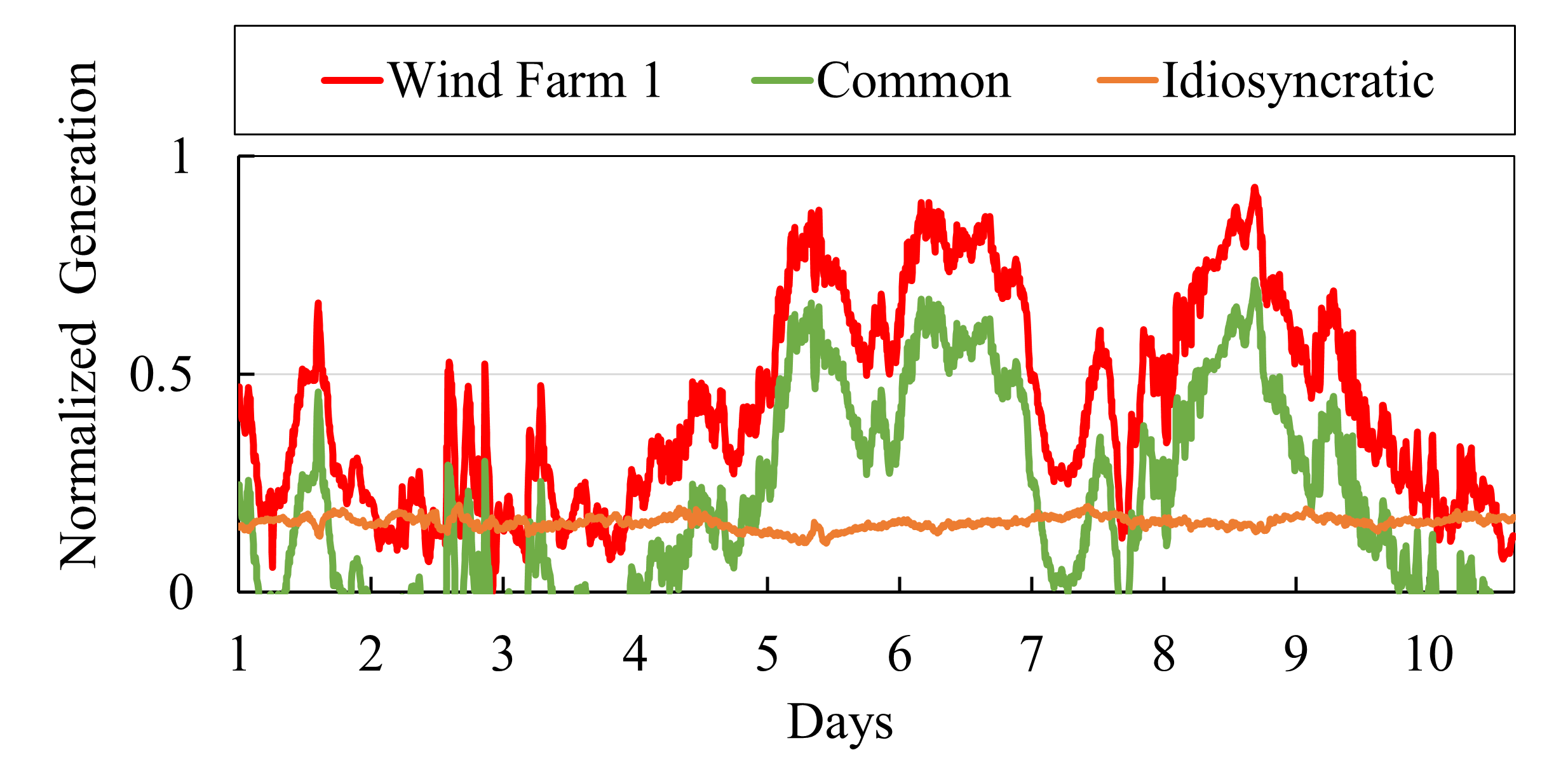}}\\
\subfloat[Wind Farm 2]{\includegraphics[scale=0.4]{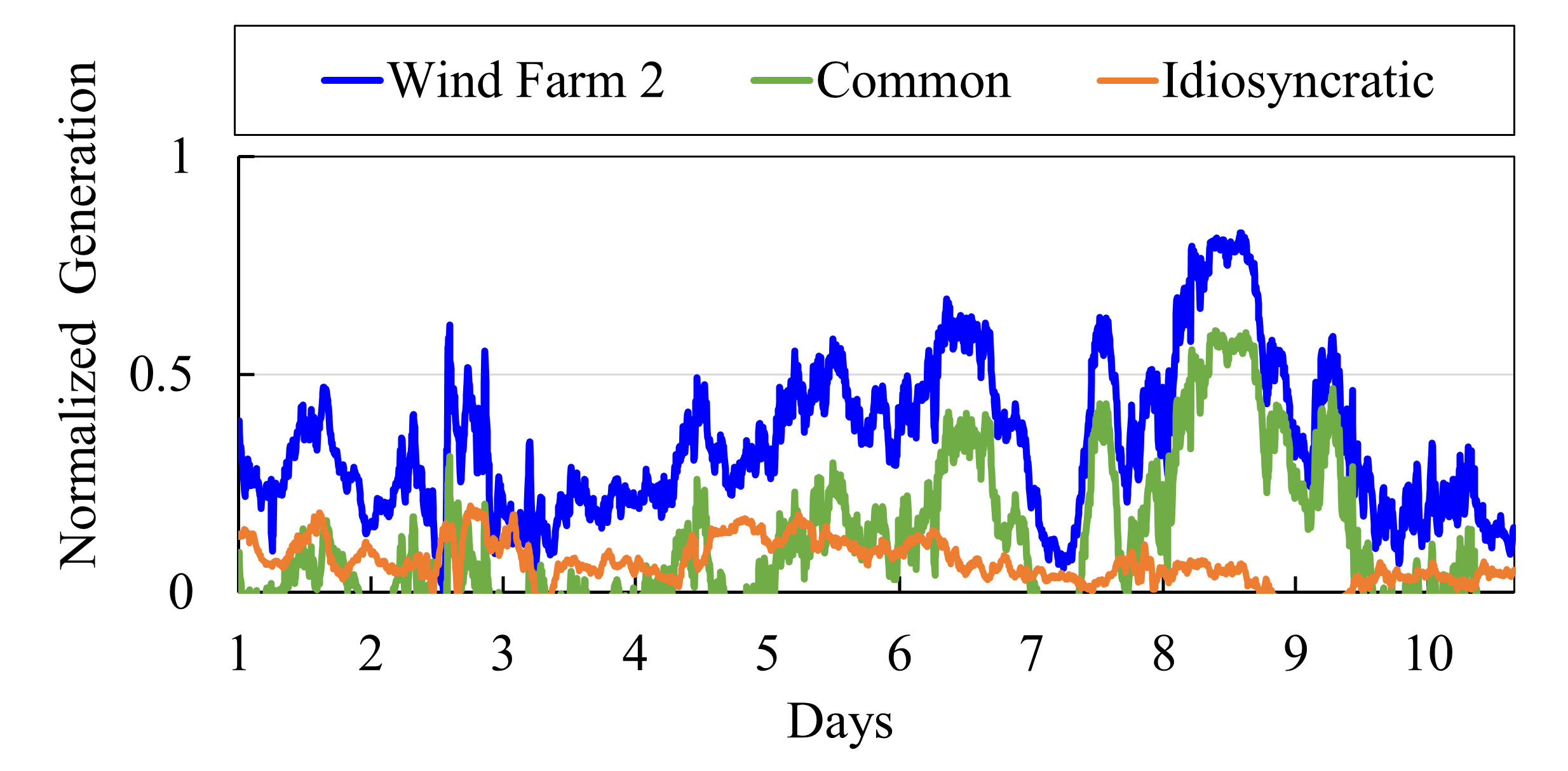}}
\caption{Comparison of the common and idiosyncratic components of the synthesized scenarios through the GDFM+GAN model.} \label{scenario_common_idio}
\end{center}
\end{figure}

\begin{figure}[t!]
\begin{center}
\centering
\subfloat[Actual wind power outputs in 2017]{\includegraphics[scale=0.4]{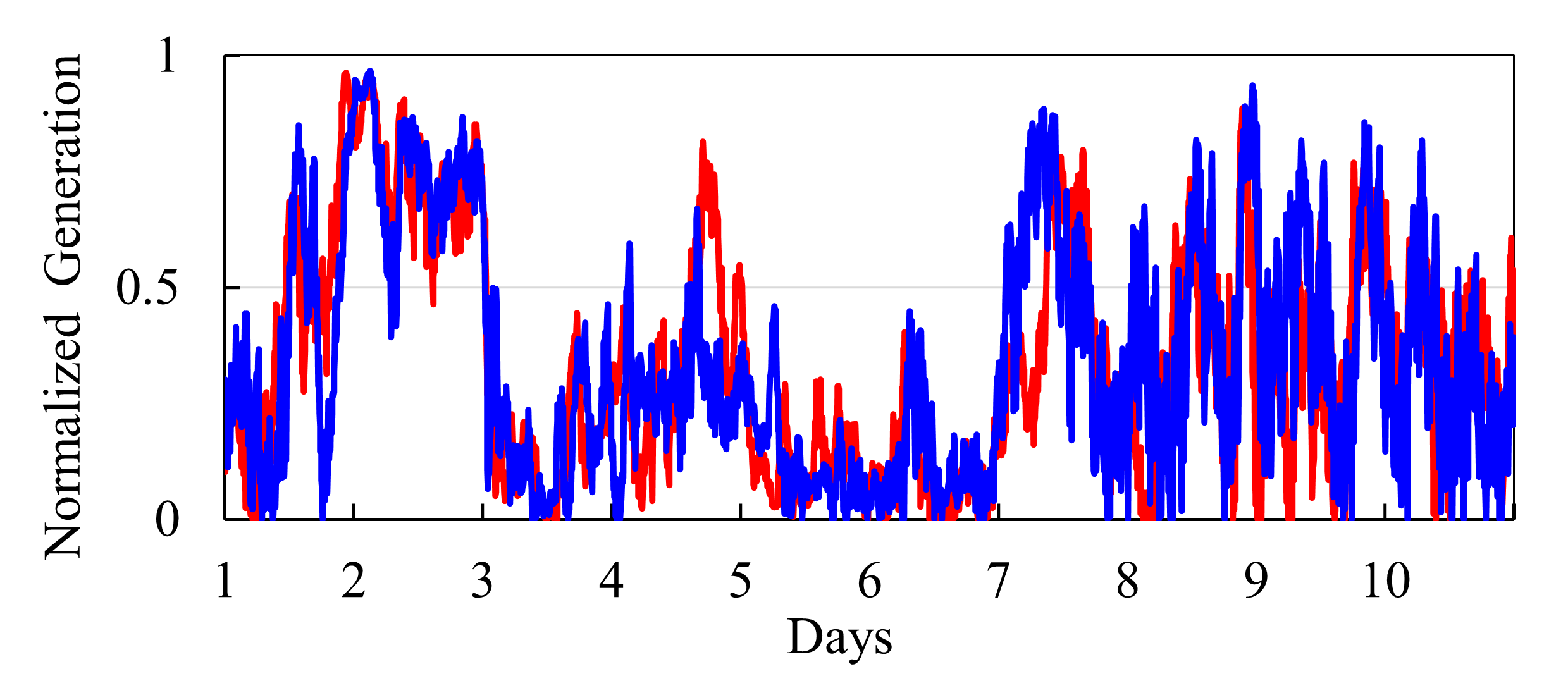}}\\ 
\subfloat[GDFM]{\includegraphics[scale=0.4]{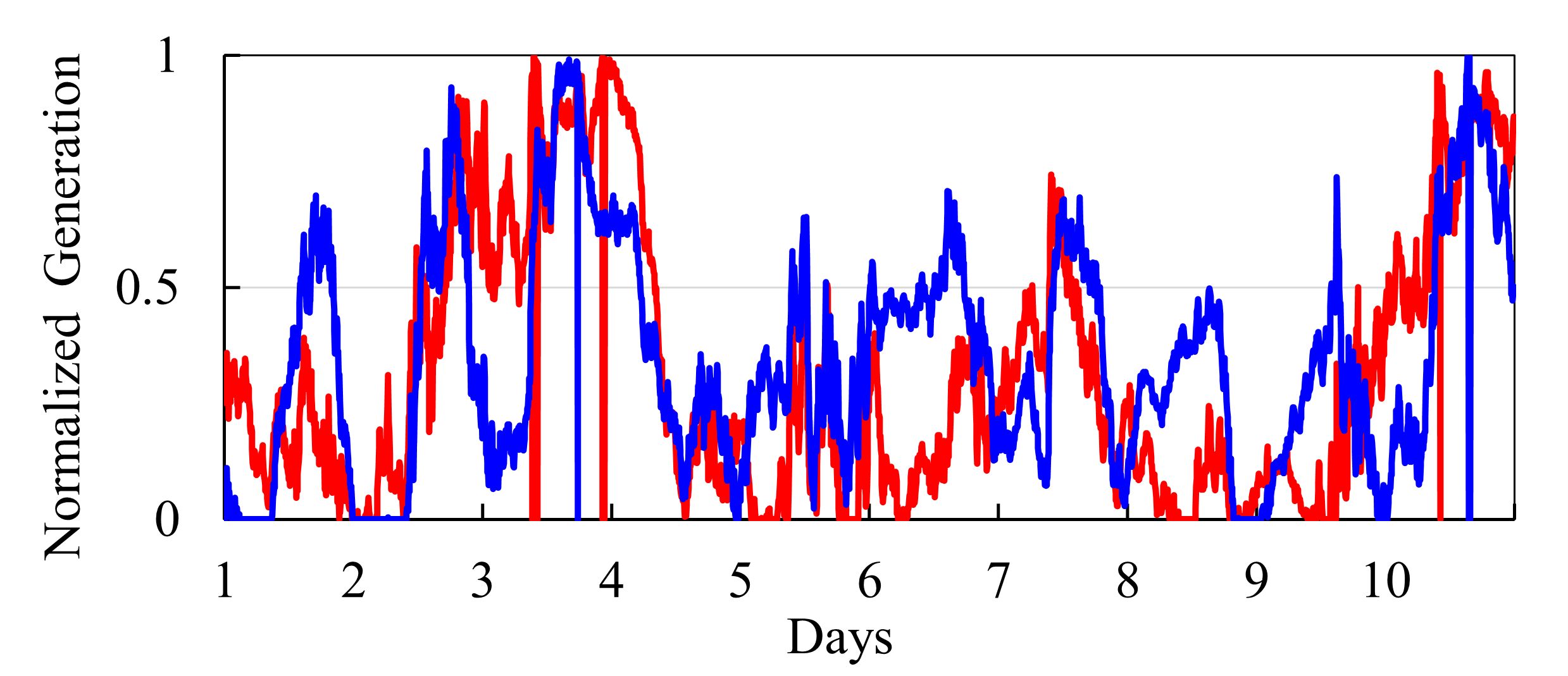}}\\
\subfloat[GAN]{\includegraphics[scale=0.4]{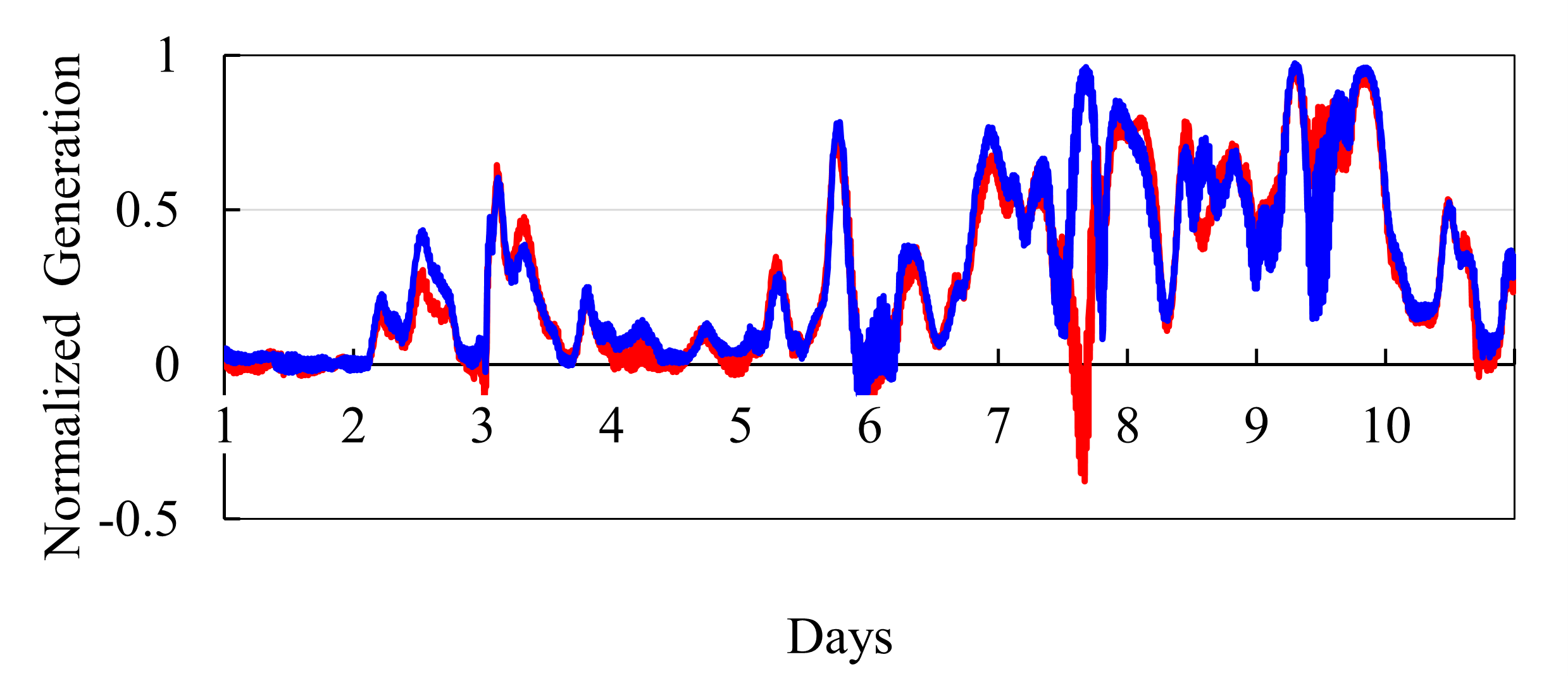}}\\
\subfloat[GDFM+GAN]{\includegraphics[scale=0.4]{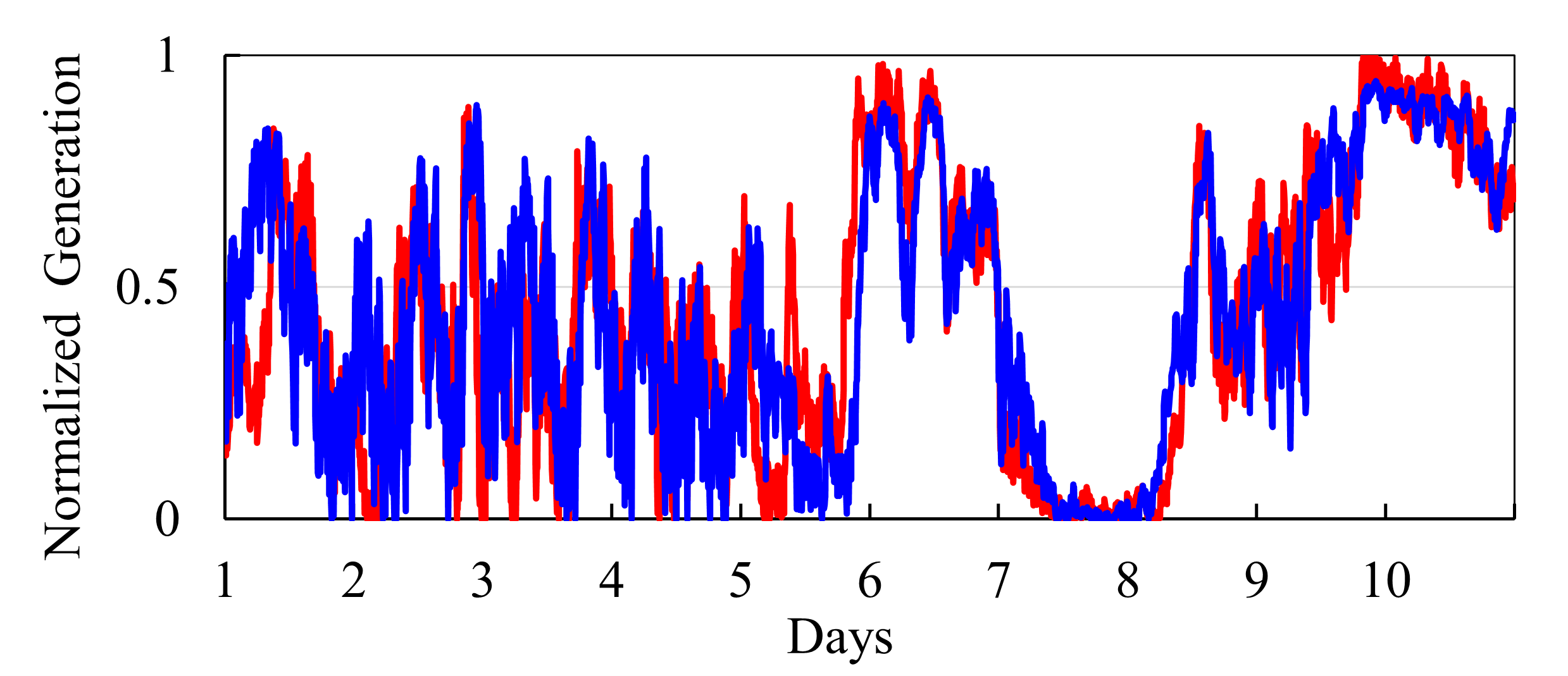}}
\caption{Comparison of 10-day wind power outputs under extreme conditions from two adjacent Australian wind farms: (a) actual measurements and scenarios synthesized using (b) GDFM, (c) GAN, and (d) GDFM+GAN.} \label{scenario_ext}
\end{center}
\end{figure}

\subsection{Waveform Validation}

We verify the effectiveness of the synthesized scenarios by comparing them with actual outputs from two adjacent wind farms over 10 consecutive days during normal operating conditions (Fig.~\ref{scenario}(a)). Actual scenarios from these wind farms exhibit similar peak patterns with slight temporal shifts. Additionally, Wind Farm 2 generally produces smaller generation magnitudes compared to Wind Farm 1.
Scenarios synthesized solely by the GDFM are shown in Fig.~\ref{scenario}(b). Due to the GDFM's explicit modeling of common and idiosyncratic components in the frequency domain, these scenarios successfully capture the overall variability of the actual data. However, the GDFM alone struggles to reproduce marginal distributions and complex waveform variations accurately, leading to insufficient representation of temporal correlations such as realistic peak shifts between adjacent wind farms.
Conversely, the scenarios generated solely by the GAN, shown in Fig.~\ref{scenario}(c), demonstrate GAN's strength in synthesizing realistic waveform shapes and marginal distributions directly from time-domain data. However, GAN-based methods typically fail to adequately represent frequency-domain correlations when applied directly in the time domain, resulting in nearly identical peaks occurring simultaneously at adjacent wind farms. Thus, the GAN alone inadequately captures realistic spatial correlation patterns.

However, the scenarios synthesized by our proposed GDFM+GAN model (Fig.~\ref{scenario}(d)) effectively address these limitations. By synthesizing the dynamic filter in the frequency domain with GAN and integrating it into the GDFM, our approach leverages the complementary advantages of each component model while mitigating their individual shortcomings. Specifically, as illustrated clearly in Fig.~\ref{scenario_common_idio}, the GDFM explicitly extracts the common component representing concurrent dynamic behaviors, such as prevailing wind patterns driven by global meteorological conditions that similarly affect multiple wind farms but with varying timing and intensity. Simultaneously, it isolates an idiosyncratic component capturing farm-specific random fluctuations driven by local conditions. Integrating these frequency-domain components into the GAN-based dynamic filter allows our model to accurately preserve realistic spectral densities, spatial-temporal correlations, and ramping characteristics. Consequently, synthesized scenarios from our combined model closely reflect actual wind power behaviors, significantly outperforming scenarios produced by either the GDFM or the GAN alone.

To further validate our approach, we examine an extreme condition scenario over a 10-day period, illustrated in Fig.~\ref{scenario_ext}(a). This period is characterized by unusually high variability and steep ramping events, which can lead to grid instability and thus warrant special attention in long-term reliability studies. Such extremes rarely occur in historical data, making it critical for scenario generation algorithms to accurately replicate these conditions for future planning.
In Fig.\ref{scenario_ext}(b), the scenario synthesized solely by the GDFM under extreme conditions exhibits large-scale ramping. However, its temporal correlations are weak, and the resulting waveform appears distorted, so this would lead to an underestimation of extreme ramp rates. In contrast, the scenario that is generated solely by the GAN in Fig.\ref{scenario_ext}(c), reproduces marginal distributions well. However, because of insufficient frequency-domain information, it fails to capture the enhanced ramp characteristics, so it results in a waveform that still resembles a normal condition despite the underlying high variability.

Our proposed GDFM+GAN method in Fig.~\ref{scenario_ext}(d) overcomes the limitations observed in the standalone models by effectively capturing high ramping events with accurate temporal correlations. By integrating a GAN-based dynamic filter into the GDFM, our approach preserves the spectral densities and spatial-temporal structure inherent in extreme conditions. The GDFM’s frequency-domain decomposition effectively isolates global meteorological drivers and local fluctuations, while the GAN refines the waveform details and marginal distributions. As a result, the synthesized scenarios could closely resemble the original extreme cases, and they could accurately model steep ramp events and essential temporal offsets needed for reliable stability analyses.

\begin{figure}[t!]
\centering
\subfloat[Actual data]{\includegraphics[scale=0.16]{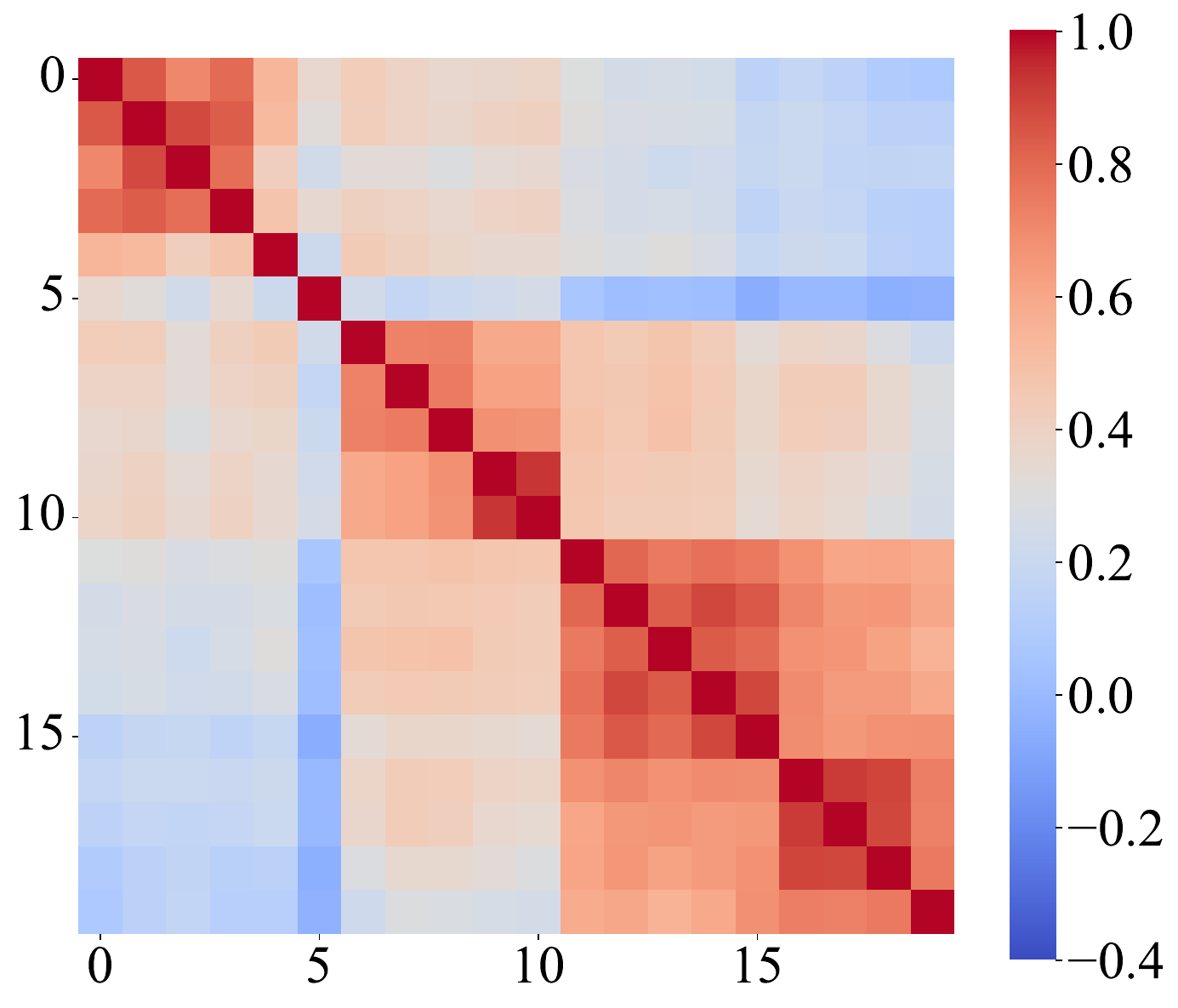}\label{Cov_AU_act}}
\quad
\subfloat[GDFM]{\includegraphics[scale=0.16]{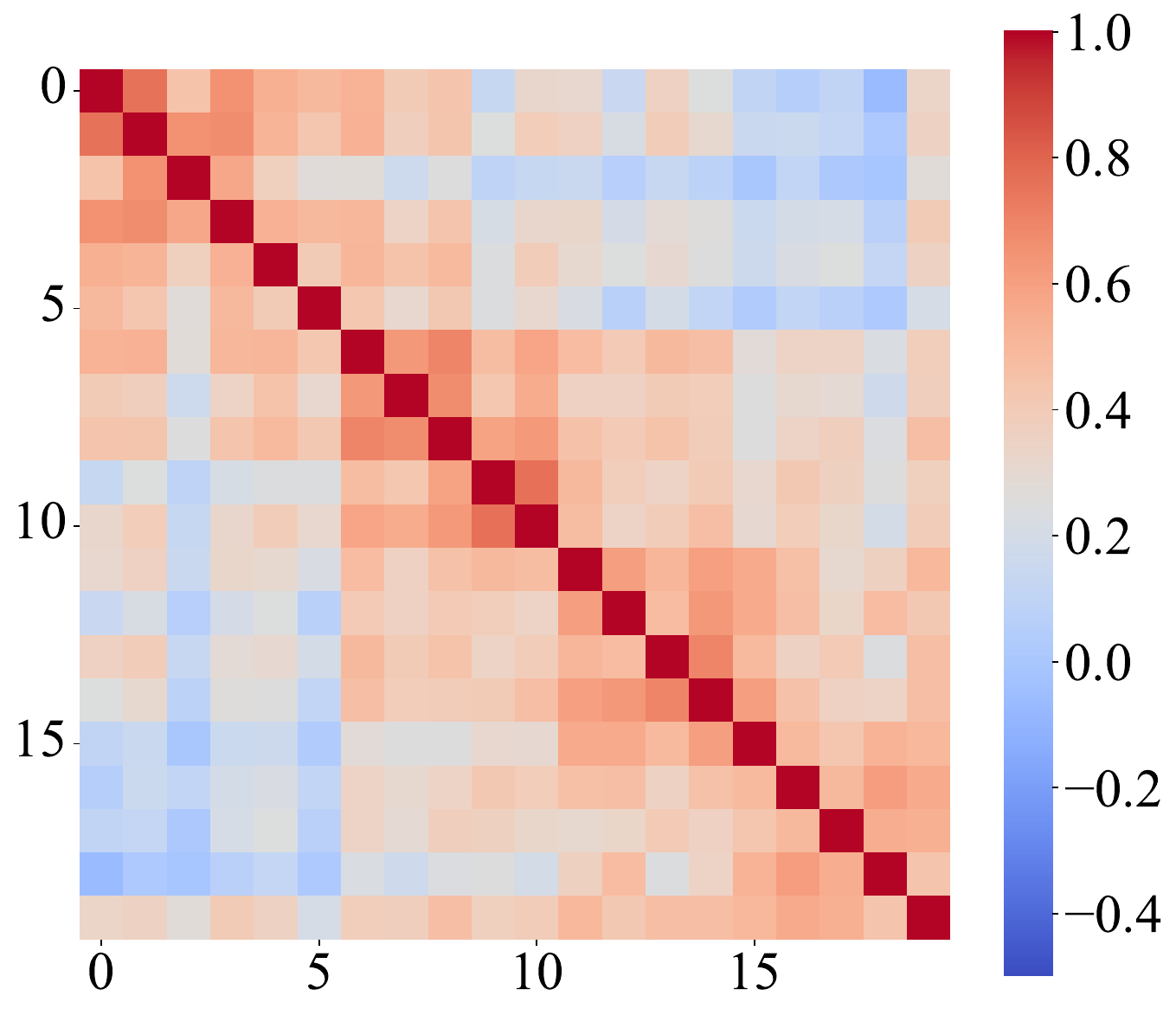}\label{Cov_AU_act_R}}\\
\subfloat[GAN]{\includegraphics[scale=0.16]{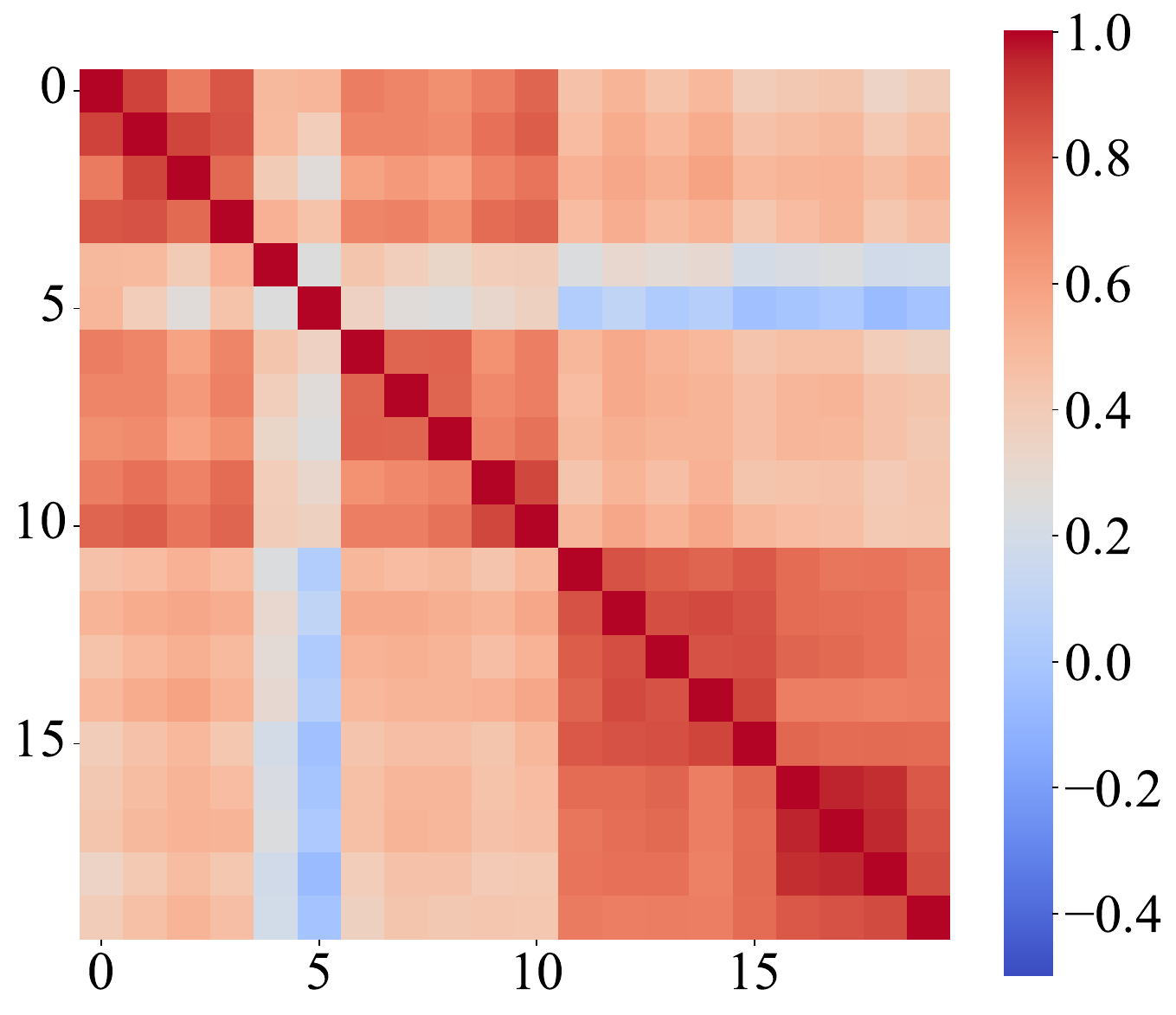}\label{Cov_AU_gen_C}}
\quad
\subfloat[GDFM+GAN]{\includegraphics[scale=0.16]{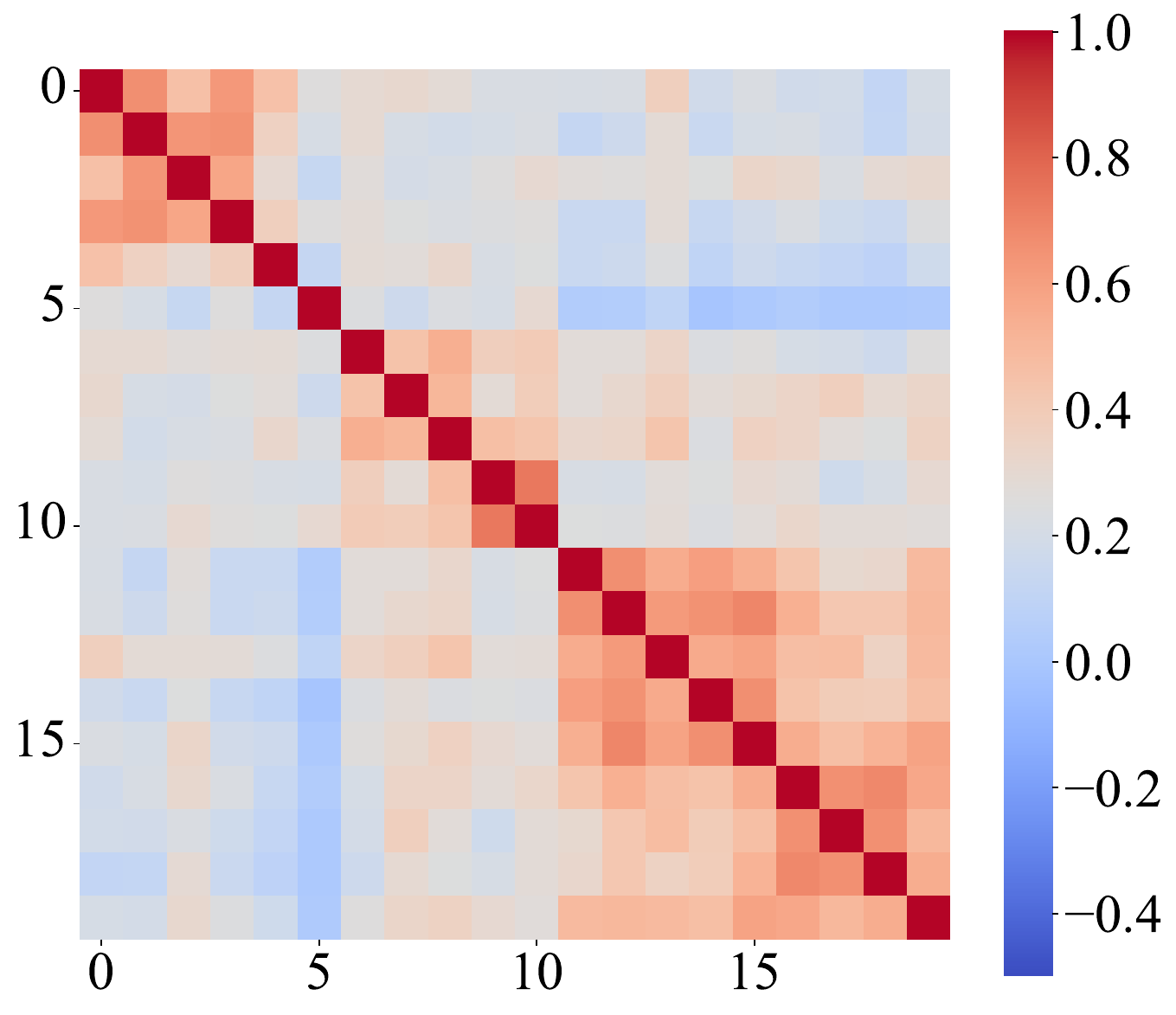}\label{Cov_AU_gen}}
\caption[]{Comparison of the covariance of the actual and synthesized scenarios.}\label{Cov}
\end{figure}

\subsection{Spatial Correlation Validation}

We verify the spatial correlation of the synthesized scenarios by comparing covariance matrices across different wind farms in Fig.~\ref{Cov}.
In detail, Fig.~\ref{Cov}(a) displays the covariance of the actual wind farm scenarios, while Figs.~\ref{Cov}(b)-(d) illustrate the covariances obtained from scenarios synthesized by the GDFM, the GAN, and our proposed GDFM+GAN, respectively. In each covariance heatmap, colors closer to red indicate stronger positive correlations.
Recall that the GDFM explicitly models spatial relationships via its dynamic filters. These filters correspond directly to the common component, which captures concurrent dynamic behaviors such as prevailing wind patterns across multiple farms. Because of this explicit representation, scenarios synthesized by the GDFM preserve the key spatial correlation features observed in the actual data. In contrast, scenarios generated solely by the GAN show much weaker agreement with spatial correlations, as GANs lack a built-in mechanism for modeling spatial dependence and rely entirely on implicit learning from raw time-series data.
This contrast underscores the importance of explicitly representing spatial correlations through the GDFM dynamic filter. By combining the GDFM with the GAN, our proposed GDFM+GAN model leverages both explicit spatial modeling and flexible waveform synthesis to produce scenarios that closely match the spatial relationships seen in actual wind power outputs.

\begin{figure}[t!]
\centering
\subfloat[]{\includegraphics[scale=0.20]{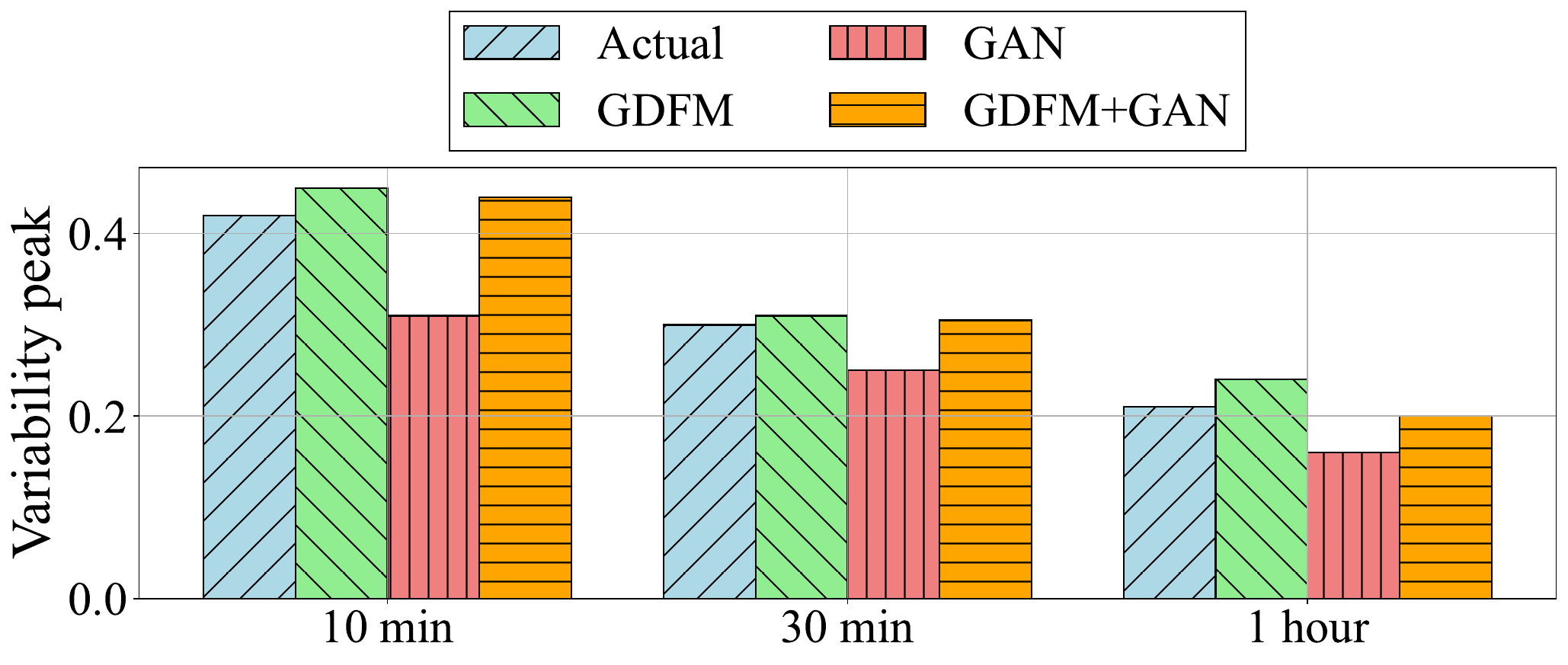}\label{Var_peak}}\\
\vspace{0.1em}
\subfloat[]{\includegraphics[scale=0.20]{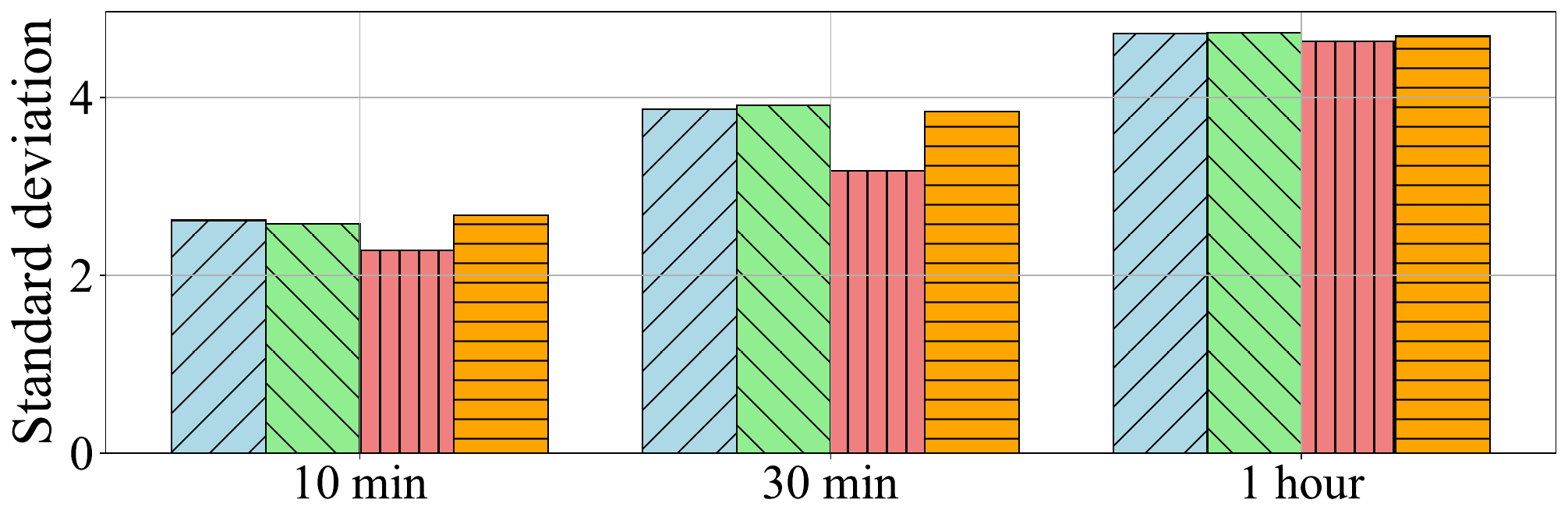}\label{Var_std}}\\
\caption[]{Comparison of the (a) peak value and (b) standard deviation of the actual and synthesized scenarios for various time intervals.}\label{Var}
\end{figure}

\begin{figure}[t!]
\centering
\subfloat[10min variability]{\includegraphics[scale=0.18]{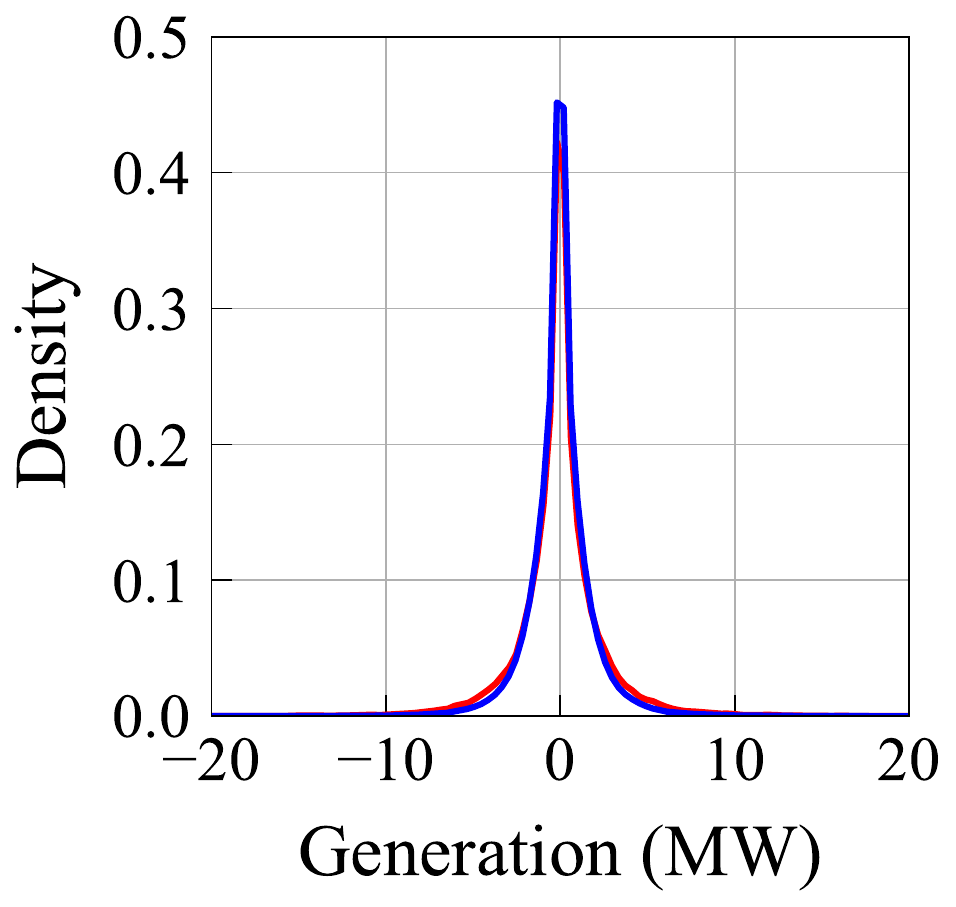}\label{Var_AU_10min}}
\subfloat[30min variability]{\includegraphics[scale=0.18]{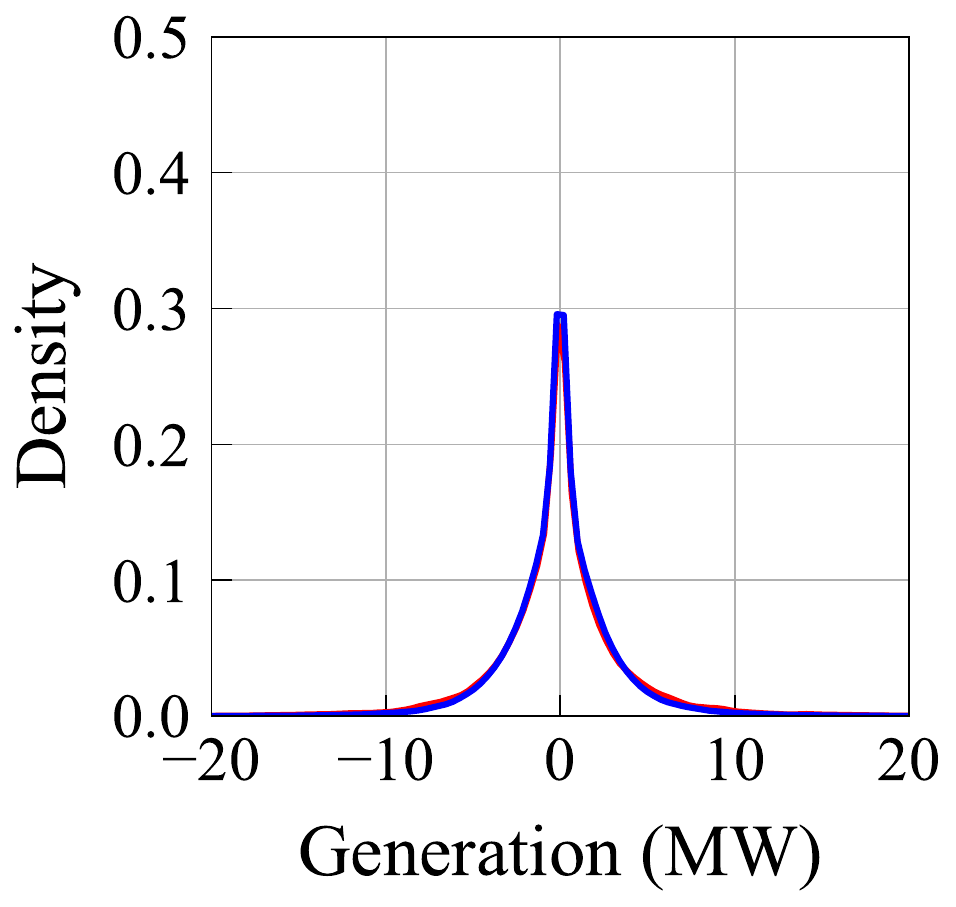}\label{Var_AU_30min}}
\subfloat[1hour variability]{\includegraphics[scale=0.18]{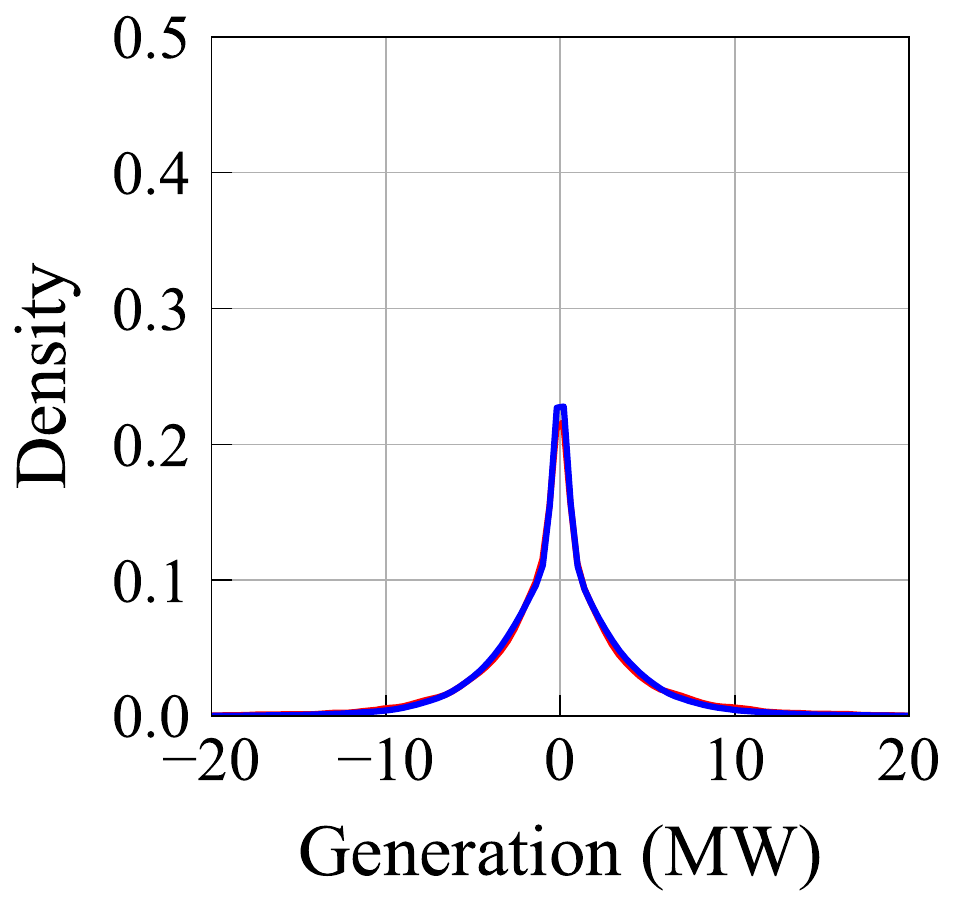}\label{Var_AU_1h}}\\
\subfloat{\includegraphics[scale=0.3]{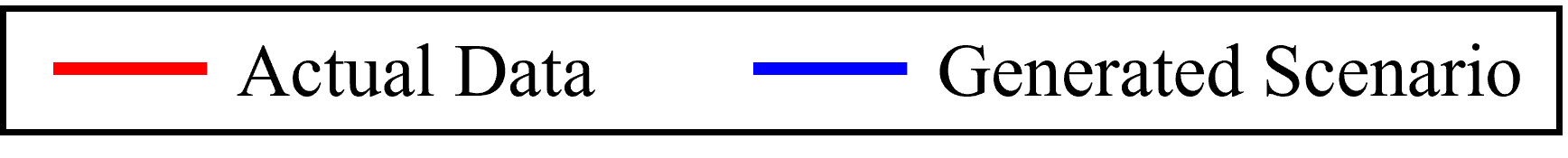}}
\hfill
\caption[]{Comparison of the variability of the actual and synthesized scenario by the GDFM+GAN for various time intervals}\label{Var_comb}
\end{figure}

\subsection{Variability Validation}
We verify that synthesized scenarios have similar variability to actual wind power.
We can compare variabilities directly by comparing the Laplace distribution under the assumption that histograms follow the Laplace distribution.
In Fig.~\ref{Var}, the peak value and standard deviation (SD) of ramping rates distributions of actual and synthesized scenarios at several intervals are represented in bar graphs.
The GDFM alone closely reproduces both peaks and SDs, thanks to its frequency domain factor loading derived from the CPSD. In contrast, the GAN alone model consistently underestimates both measures because it lacks frequency‑based variability modeling.
By embedding the GAN‑synthesized dynamic filter into the GDFM, our combined GDFM+GAN approach preserves the GDFM’s precise variability representation. This close agreement between the actual and synthesized distributions for the GDFM+GAN is highlighted in Fig.~\ref{Var_comb}, demonstrating the advantage of combining explicit variability modeling in the frequency domain.

\begin{table}[t!]
\caption{The variability of the actual and generated scenarios through the proposed GDFM+GAN model.}
\begin{center}
\begin{tabular}{c|c|c|c}
\Xhline{3\arrayrulewidth}
Data  & 10min & 30min & 1hour \\
\hline
\makecell{Distribution parameter\\of actual scenario ($\mu$, $b$)} & \makecell{(1.3747e-04,\\1.2736e-02)}  &\makecell{(3.8356e-04,\\1.2541e-02)}  & \makecell{(5.3977e-04,\\1.2413e-02)}  \\
\hline
\makecell{Distribution parameter\\of synthesized scenario} &\makecell{(1.2432e-04,\\1.2743e-02)}  & \makecell{(2.9010e-04,\\1.2469e-02)} & \makecell{(3.8807e-04,\\1.2325e-02)} \\
\hline
KL divergence & 0.2617 & 0.2400 & 0.2277\\
\Xhline{3\arrayrulewidth}
\end{tabular} \label{Var_Table}
\end{center}
\end{table}

We can also compare the Kullback-Leibler (KL) divergence of the actual and synthesized scenarios' variability distributions. The KL divergence $D_{KL}(P||Q)$ is defined as
\begin{align}
D_{KL}(P||Q)=\sum_{x \in \chi}P(x)\ln \left(\frac{P(x)}{Q(x)}\right), \label{KL}
\end{align}
which measures the difference between the reference distribution $P$ and the other distribution $Q$. If the KL divergence is close to zero, two distributions are close to equivalent. In Table \ref{Var_Table}, the distribution parameters and the KL divergences for three different intervals of the synthesized data through the GDFM+GAN are described. The KL divergences of 10, 30 minutes, and one hour are less than 0.3, so we can conclude that the variability of synthesized scenarios is similar to that of actual scenarios. 
Finally, our synthesized scenarios have similar variability to the actual scenario. 

\begin{figure}[t!]
\centering
\subfloat[GDFM]{\includegraphics[scale=0.25]{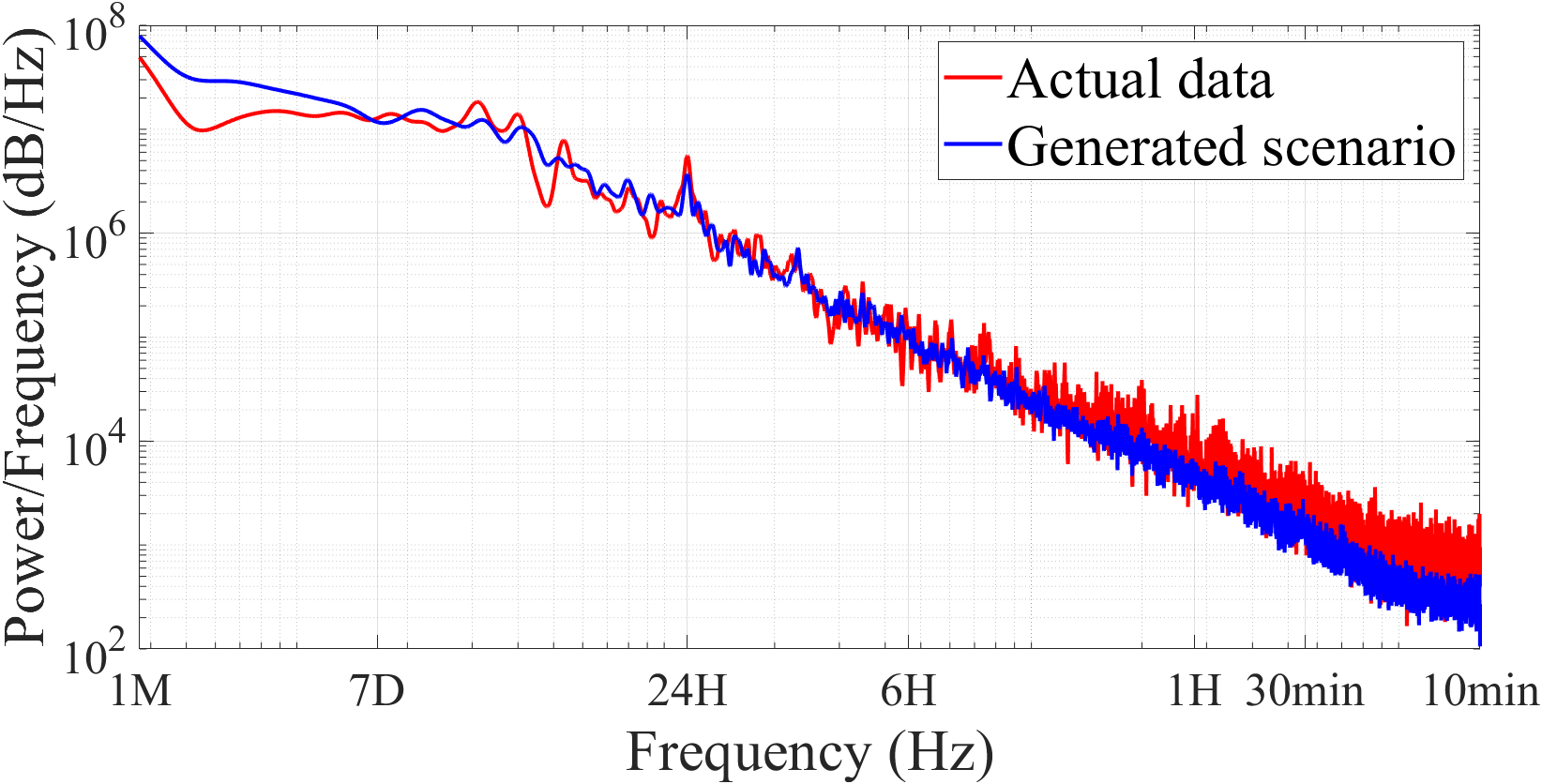}\label{PSD_R}}\\
\subfloat[GAN]{\includegraphics[scale=0.25]{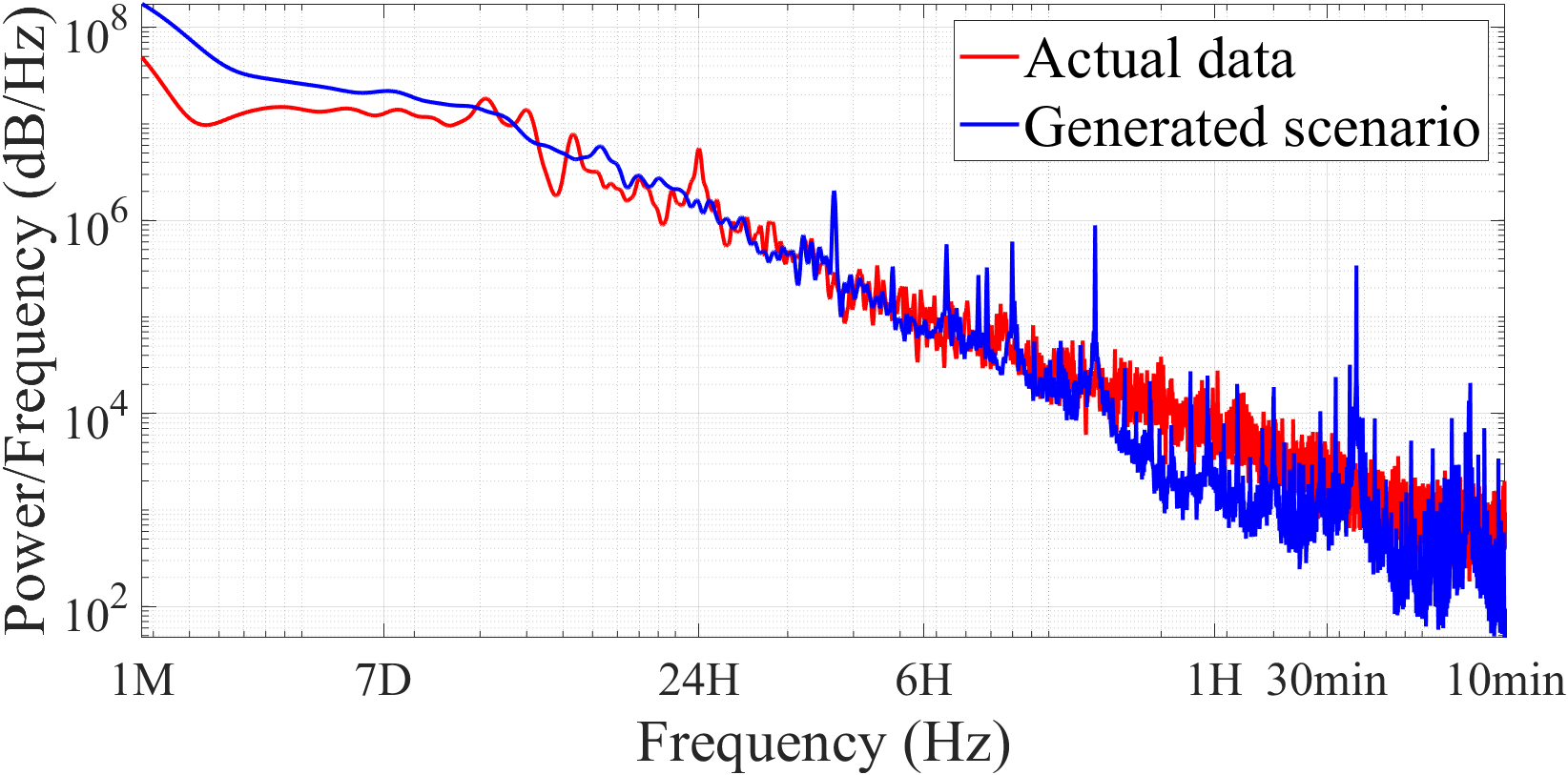}\label{PSD_C}}\\
\subfloat[GDFM+GAN]{\includegraphics[scale=0.25]{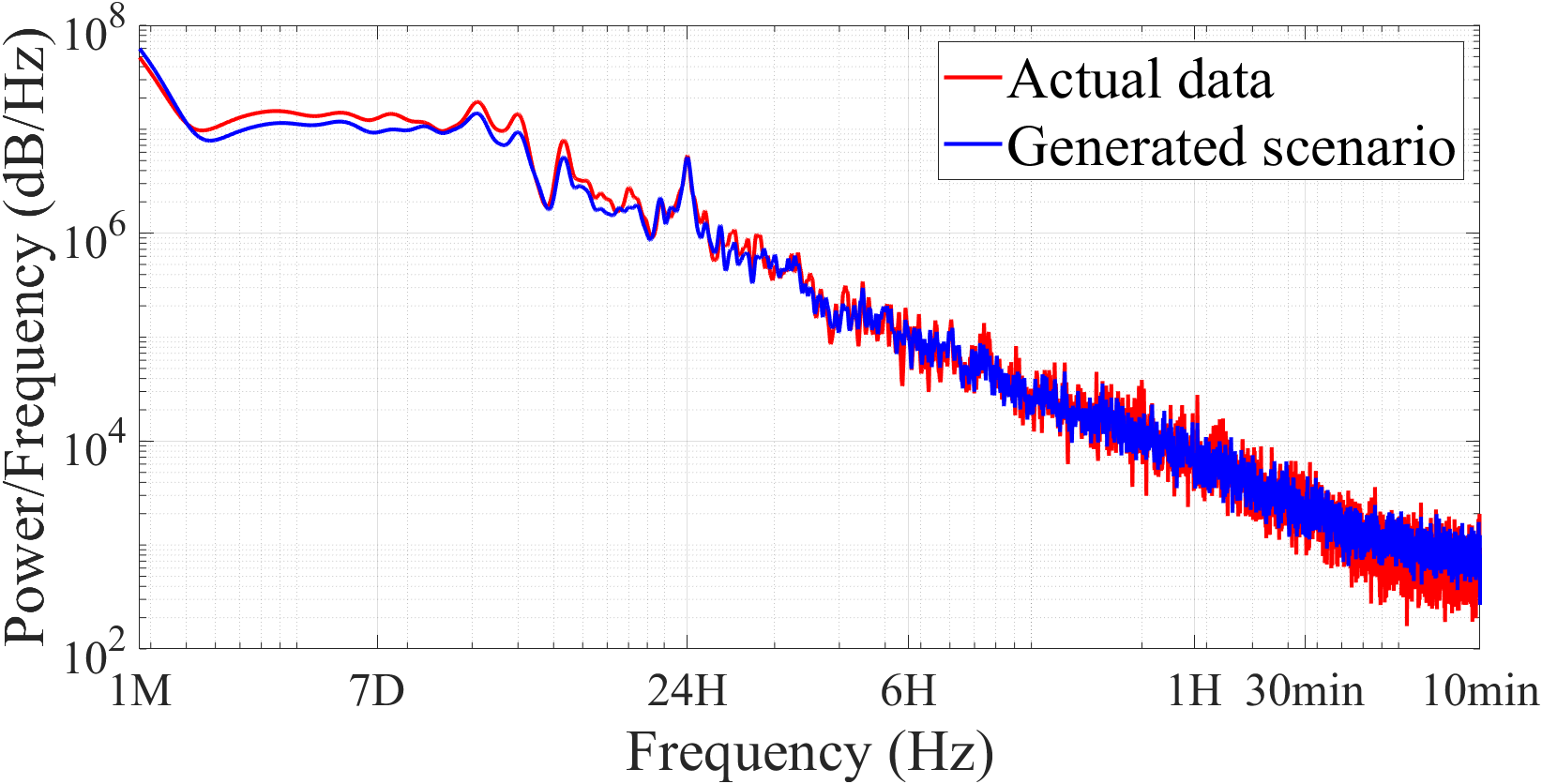}\label{PSD_G}}
\caption[]{Comparison of the PSDs of the actual and synthesized scenarios.}\label{PSD}
\end{figure}

\subsection{Power Spectral Density Validation}
We compare the PSDs of synthesized scenarios of individual wind farms to those of actual scenarios.
A similar PSD can show that the synthesized scenarios have similar stochastic characteristics as an actual scenario since the magnitude of the PSD usually represents the magnitude of ramping events. In Fig. \ref{PSD}, PSDs of actual and synthesized scenarios are estimated by Welch's method.
%

We can observe several interesting results. First, the PSDs of the scenario synthesized by the GDFM in Fig.~\ref{PSD_R} and GDFM+GAN in Fig.~\ref{PSD_G} are similar to the actual PSD, but fluctuation in high frequencies corresponding to periods shorter than 20 minutes slightly differs from fluctuations of the actual PSD. 
The reason might be in the phase angle $\angle\bm B(\omega_m)$ and $\bm A({\omega _m})$. When we synthesize scenarios by changing $\bm B(\omega_m)$, we synthesize the magnitude of $\bm B(\omega_m)$ using $\hat{\bm G}(z)$, but we use the observed angle $\bm B(\omega_m)$. If $\hat{\bm G}(z)$ is estimated, $\angle\bm B(\omega_m)$ should be fit to $\hat{\bm G}(z)$. However, we do not have a model for the relationship between $\hat{\bm G}(z)$ and $\angle\bm B(\omega_m)$. Therefore, the discrepancy between $\hat{\bm G}(z)$ and $\angle\bm B(\omega_m)$ might cause a bigger difference in GDFM than in GDFM+GAN.

%
In addition, the PSD of the scenario synthesized by the GAN has bigger fluctuation than that of the actual PSD in Fig. \ref{PSD_C}. When we only use the GAN, the scenarios cannot follow the actual distribution. Since the GAN just imitates the waveform itself, and it does not analyze the data in the frequency domain, it cannot fully represent the spectral characteristic. Although scenarios look like actual data in the time domain, the characteristics in the frequency domain may not be the same as the actual scenario.

Furthermore, we calculate the cumulative energy generation of the PSD synthesized by the GDFM+GAN for a given frequency range in Table \ref{PSD_Table}.
The cumulative energy represents the power of variability in wind power. The difference is normalized by the cumulative energy of all frequencies.
We can observe that there is a small difference between the two energies under the PSDs between the ten-minute period and the one-hour period, which is around 0.06\% of the actual scenario. The difference in cumulative energy between seven days and one day is 5.18\%. The PSDs of the actual and synthesized scenarios across the entire frequency range do not differ significantly, which confirms the adequacy for use in resource assessments. Finally, the synthesized scenarios exhibit a power spectral density similar to that of the actual scenario.


\begin{table}[t!]
\caption{The cumulative energy of PSDs for the actual and the generated scenarios through the GDFM+GAN model.}
\begin{center}
\begin{tabular}{c|c|c|c|c}
\Xhline{3\arrayrulewidth}
Data & 7D - 1D & 1D - 6h & 6h - 1h & 1h - 10m\\
\hline
\makecell{Actual\\Data} & 50.65 dB & 13.70 dB & 5.60 dB & 2.23 dB\\
\hline
\makecell{Synthesized\\Scenario}& 41.31 dB & 12.72 dB & 5.48 dB& 2.34 dB\\
\hline
Difference & 5.18\% & 0.54\% & 0.06\% & 0.05\% \\
\Xhline{3\arrayrulewidth}
\end{tabular} \label{PSD_Table}
\end{center}
\end{table}

\subsection{Other Statistical Characteristics}
\begin{figure}[t!]
\begin{center}
\includegraphics[scale=0.25]{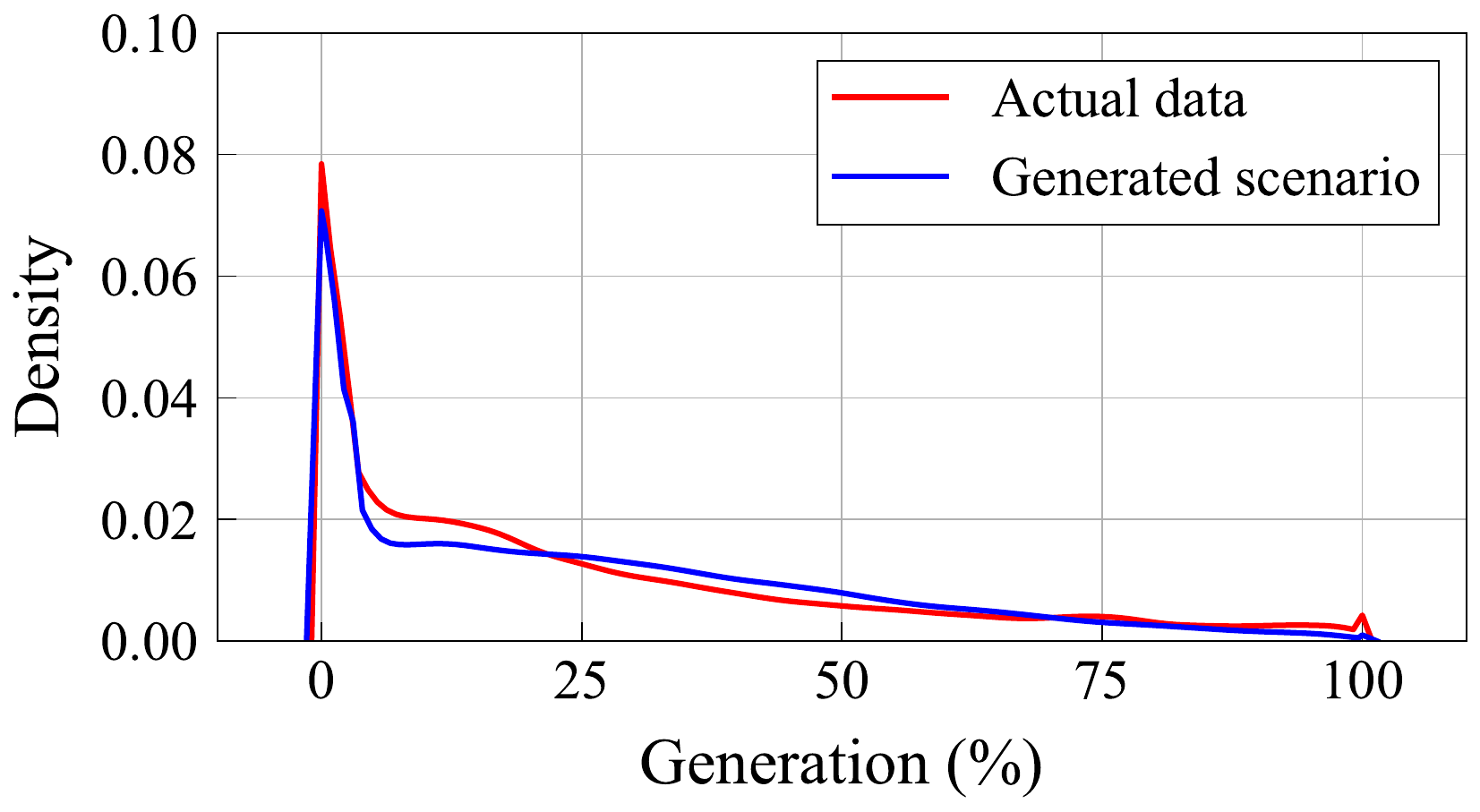}
\caption{Comparison of distributions of the actual and synthesized scenarios through the GDFM+GAN model.} \label{WPD}
\end{center}
\end{figure}

We verify that the synthesized scenarios have similar statistical characteristics to actual scenarios. We compare the mean, variance, and capacity factor (CF) of the actual and scenarios synthesized by the GDFM+GAN. The mean and variance of the synthesized scenarios are different by 4.79$\%$ and 3.02$\%$ of the actual data. Furthermore, the difference in CF between the actual and synthesized scenarios is only 3.81\%. Moreover, the marginal distributions of actual and synthesized scenarios in Fig.~\ref{WPD} are also very similar. Therefore, the actual and synthesized scenarios have similar statistical characteristics.

\section{Conclusion}
We proposed a novel wind power scenario generation method that integrates the generalized dynamic factor model (GDFM) with a generative adversarial network (GAN). The GDFM explicitly models common and idiosyncratic components in the frequency domain, thereby capturing spatial and frequency correlations and the smoothing effect from wind farm aggregation, though it struggles with marginal distributions and fine waveform details. In contrast, GANs excel at reproducing realistic waveforms and marginal distributions from time-domain data but often fail to preserve frequency-domain correlations. Our approach uses a GAN to synthesize a dynamic filter that is embedded into the GDFM, effectively combining the strengths of both models. As a result, the synthesized scenarios accurately replicate marginal distributions, detailed waveforms, realistic ramping events, spectral densities, and spatial-temporal correlations observed in actual wind power data—thereby enhancing the realism and reliability of scenarios for operational planning.

Future work will explore integrating alternative conditional generative models—such as variational autoencoders (VAEs) and diffusion models—with the GDFM framework. Although these models require an explicit latent prior and incur higher computational costs, they offer improved training stability and mode coverage. We anticipate that their integration will further enhance wind power scenario synthesis. Additionally, future studies will extend our analysis to include diverse geographic regions and wind farms of various scales (e.g., onshore and offshore), to account for the significant variations in wind patterns observed in more geographically dispersed installations.

\ifCLASSOPTIONcaptionsoff
\newpage
\fi

\bibliographystyle{IEEEtran}
\bibliography{DCGAN_ref.bib}

\end{document}